\PassOptionsToPackage{table}{xcolor}
\documentclass{eitnlp}

\usepackage{amsmath}
\usepackage{amssymb}
\usepackage{array}
\usepackage{booktabs}
\usepackage{graphicx}
\usepackage{makecell}
\usepackage{microtype}
\usepackage{multirow}
\usepackage{placeins}
\usepackage{hyperref}
\usepackage{url}
\usepackage{xspace}
\usepackage{makecell}
\usepackage{enumitem}       
\usepackage{array}          

\usepackage{float}
\usepackage{wrapfig}
\usepackage{algorithm}
\usepackage{algpseudocode}

\definecolor{oursblue}{RGB}{226,237,247}
\definecolor{oursteal}{RGB}{226,242,239}
\definecolor{lightgray}{RGB}{245,246,247}

\newcommand{\best}[1]{\textbf{#1}}
\newcommand{\second}[1]{\underline{#1}}

\algrenewcommand{\algorithmicrequire}{\textbf{Input:}}
\algrenewcommand{\algorithmicensure}{\textbf{Output:}}
\algrenewcommand{\algorithmicindent}{1em}
\algrenewcommand{\alglinenumber}[1]{\scriptsize #1:}

\title{GeoGR: Spatial–Semantic Token Compression and Group Routing for Efficient 3D VLMs}
\runningtitle{GeoGR}
\author{\textbf{Yongxin Zheng}\textsuperscript{1,*} \enspace
        \textbf{Hao Wu}\textsuperscript{1,2,*} \enspace
        \textbf{Fang Li}\textsuperscript{1} \enspace
        \textbf{Hu Fu}\textsuperscript{1} \enspace
        \textbf{Peiran Yin}\textsuperscript{1,2} \enspace
        \textbf{Xudong Wang}\textsuperscript{1} \\
        \textbf{Haozhe Hu}\textsuperscript{1,2} \enspace 
        \textbf{Xinghao Chen}\textsuperscript{1,3} \enspace 
        \textbf{Yunpu Ma}\textsuperscript{4} \enspace 
        \textbf{Wei Zhang}\textsuperscript{1} \enspace 
        \textbf{Xiaoyu Shen}\textsuperscript{1,\textdagger} \\
}
\affiliation{%
  \textsuperscript{1}EIT-NLP Lab, Eastern Institute of Technology, Ningbo \\
  \textsuperscript{2}Shanghai Jiao Tong University \qquad \textsuperscript{3}The Hong Kong Polytechnic University \\ 
  \textsuperscript{4}Munich Center for Machine Learning, LMU Munich\\
}
\authornotes{\textsuperscript{*}Equal contribution.\qquad
             \textsuperscript{\textdagger}Corresponding author.
}
\contact{\texttt{haowu.83@sjtu.edu.cn \quad xyshen@eitech.edu.cn}
}

\begin{document}
\maketitle

\begin{abstract}
Multi-view representations enable pretrained image and video vision-language models to understand 3D scenes, but their long visual token sequences make inference costly. Redundancy arises both from repeated observations of the same scene regions across views and from processing all visual tokens at every LLM layer, regardless of their relevance to the current computation. We propose GeoGR, a two-stage framework that jointly compresses scene representations and visual computation. At the projector level, GeoSemZip consolidates cross-view observations into 3D voxel tokens and combines semantic saliency with spatial coverage to construct a compact scene representation. Unselected voxel features are aggregated into contextual anchors to retain scene information that would otherwise be discarded. At the LLM level, we identify an effective visual computation window and observe that query-conditioned token relevance remains locally stable but changes over longer layer intervals. Guided by these findings, GroupRoute combines late entry and early exit with recoverable group-wise skipping. Low-relevance tokens bypass complete LLM layers while retaining their hidden states and can rejoin the active group at subsequent anchor layers selected under a computation budget. For compression-aware post-training, we construct a compact dataset of 67K samples by retaining 30\% of the original training pool. With this protocol, GeoGR retains 95.2\% and 96.8\% of the dense post-trained models' performance on LLaVA-OneVision-7B and Video-3D LLM-7B, respectively, while accelerating LLM prefill by $6.8\times$ and $6.1\times$.  These results demonstrate the consistent effectiveness of GeoGR across 3D VLMs with and without explicit 3D spatial encoding, highlighting a favorable performance-efficiency trade-off. The code is available at \url{https://github.com/EIT-NLP/GeoGR}.

\end{abstract}

\section{Introduction}
\label{sec:introduction}
Enabling multimodal large language models to understand 3D scenes is an important step toward spatial intelligence~\citep{hong2023threedllm}. Due to the scarcity of 3D-language data compared with large-scale image-text corpora~\citep{zhu2025llava3d, tang2025more}, recent studies often represent 3D scenes as sequences of posed RGB or RGB-D views, allowing pretrained vision-language models (VLMs) to transfer knowledge from 2D images and videos to 3D reasoning~\citep{zheng2024video3dllm, zhu2025llava3d,chen2025integrating}. However, converting a 3D scene into multiple 2D views inevitably introduces a large number of visual tokens. For example, Video-3D LLM uses 32 views per scene, yielding 6,272 visual tokens~\citep{zheng2024video3dllm}. Such long visual sequences make inference substantially more expensive, making efficiency a critical challenge for multi-view 3D VLMs.

\begin{figure*}[t]
    \centering
    \includegraphics[width=.95\textwidth]{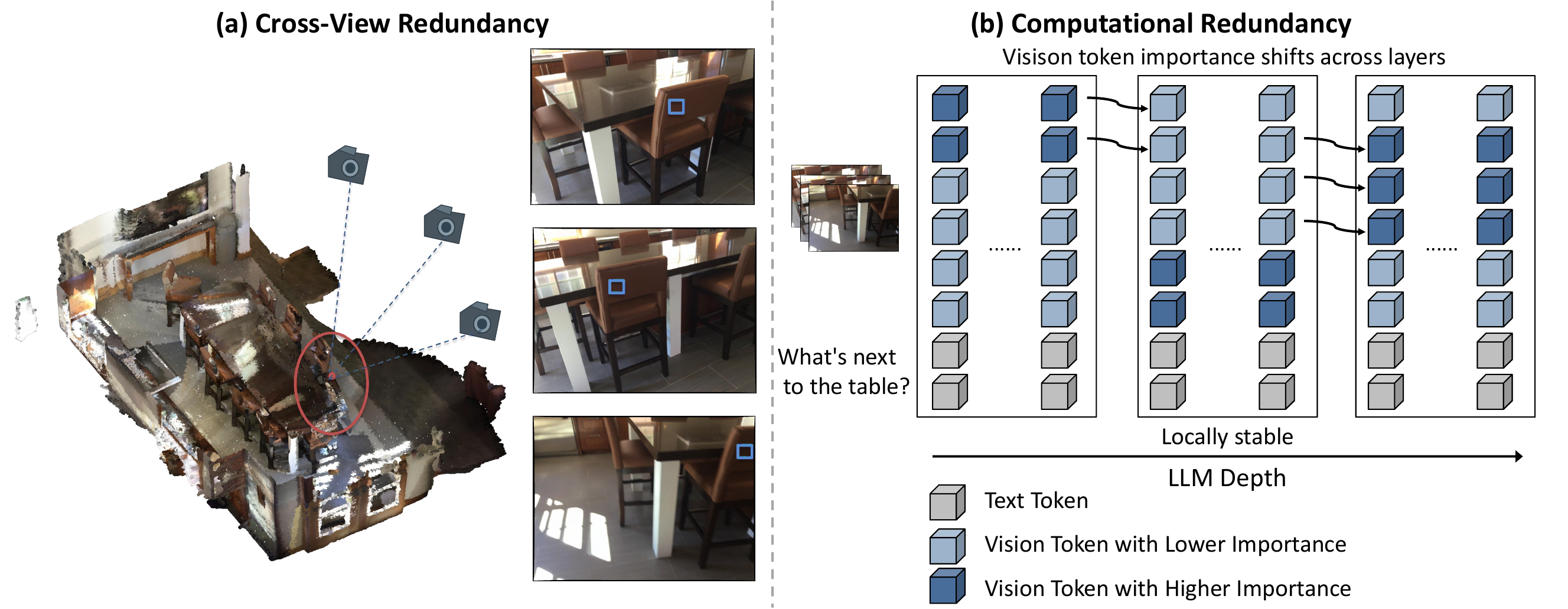}
    \caption{\textbf{Redundancy in multi-view 3D VLMs.} (a) Overlapping views repeatedly observe the same 3D region (blue boxes). (b) For the same question, visual token importance remains locally stable but shifts across more distant layers. See Section~\ref{sec:layer_dynamics} for quantitative analysis.}
    \label{fig:motivation}
    \vspace{-5pt}
\end{figure*}

Multi-view 3D VLMs exhibit redundancy in both scene representations and visual computation. At the representation level, overlapping views repeatedly encode the same physical regions (Figure~\ref{fig:motivation}(a)). While geometric correspondence allows for basic consolidation~\citep{huang2025dtc}, existing spatial pruning methods struggle to balance semantic saliency with scene coverage~\citep{li2026segpruner, ling2026coverprune, li2026geometry}. Strictly optimizing for salient objects discards the less prominent background regions essential for establishing 3D spatial context. At the computation level, dense inference processes all visual tokens at every LLM layer.
Studies of image-based VLMs suggest that the contribution of visual computation varies across layers~\citep{fan2025visipruner, fan2026what, wu2026hidrop}, motivating us to examine how visual computation should be allocated across tokens and layers in multi-view 3D VLMs. Moreover, existing pruning strategies often make irreversible execution decisions. As we observe in 3D reasoning, a token's relevance to the query shifts dynamically across layers (Figure~\ref{fig:motivation}(b)). Permanently removing tokens with low relevance at one layer may therefore discard visual evidence needed for reasoning at later layers.

To address these two forms of redundancy, we propose GeoGR, a two-stage framework that compresses multi-view scene representations and selectively allocates visual computation within the LLM. At the projector level, between the vision encoder and the LLM~\citep{wu2026data}, we introduce GeoSemZip to construct a compact scene representation. GeoSemZip maps cross-view observations into a shared 3D coordinate system and consolidates them into voxel tokens. It retains voxel tokens with high encoder-attention saliency as dominant tokens and selects contextual anchors by sampling the remaining voxels along their 3D Morton order. These anchors complement the salient tokens by covering less salient scene regions. Unselected voxel features are aggregated into the contextual anchors according to feature similarity, retaining scene information that would otherwise be discarded. GeoSemZip thus shortens the visual sequence while preserving salient content and spatial coverage.

Within the LLM, we introduce GroupRoute to reduce visual computation through Late Entry, Early Exit, and Recoverable Group-wise Skipping (RGS). Training-free interventions that separately delay visual token entry and advance visual token exit suggest a layered pattern of visual computation in multi-view 3D VLMs: portions of shallow- and deep-layer processing can be omitted with limited changes to predictions. Using next-token KL divergence relative to full visual computation, we identify candidate window boundaries, then compare the candidates after compression-aware post-training to select the final window. Within this window, query-conditioned token relevance remains locally stable but changes over longer layer intervals (Section~\ref{sec:layer_dynamics}), motivating group reuse over short intervals with periodic reassessment. At anchor layers, RGS forms high- and low-priority groups using query-conditioned attention. High-priority tokens execute subsequent LLM layers, while low-priority tokens bypass them with their hidden states preserved. All visual tokens execute the next anchor layer and are rescored, allowing previously skipped tokens to rejoin the active group. Budget-constrained Anchor and Regrouping Search (BARS) selects anchor positions to minimize the risk of reusing groups under a computation budget.

To support compression-aware post-training, we use PRISM~\citep{bi2026prism} to select the least redundant 30\% of samples from the original training pool, yielding a compact dataset of 67K samples. Using this post-training protocol, we evaluate GeoGR on LLaVA-OneVision-7B and Video-3D LLM-7B, without and with explicit 3D spatial encoding, respectively. Evaluation covers two question-answering benchmarks for LLaVA-OneVision and five benchmarks spanning question answering, visual grounding, and dense captioning for Video-3D LLM. 
At the projector level, GeoSemZip's cross-view consolidation and residual aggregation preserve salient content and scene context under a limited token budget. 
At the LLM level, GroupRoute complements this representation-level compression with query-conditioned regrouping, improving performance retention over static projector-derived grouping while keeping temporarily skipped tokens available for later reactivation.
Together, the two modules enable GeoGR to retain 95.2\% and 96.8\% of the respective dense post-trained models' performance, with LLM prefill speedups of $6.8\times$ and $6.1\times$ and time-to-first-token (TTFT) speedups of $2.4\times$ and $2.3\times$, respectively.

Our main contributions are as follows:
\textbf{(1)} We propose GeoSemZip, a projector-level compression module that combines cross-view geometric consolidation with semantic saliency and spatial coverage. It aggregates unselected voxel features into contextual anchors to retain scene information under compression.
\textbf{(2)} We identify effective visual computation windows in multi-view 3D VLMs and observe locally stable but evolving query-conditioned token relevance across layers. Guided by these findings, we propose GroupRoute, which combines Late Entry and Early Exit with Recoverable Group-wise Skipping. 
\textbf{(3)} We construct a compact 67K-sample dataset for compression-aware post-training of 3D VLMs and will release it publicly.
\textbf{(4)} We propose GeoGR, a two-stage compression framework comprising GeoSemZip and GroupRoute that achieves a favorable performance-efficiency trade-off on LLaVA-OneVision-7B and Video-3D LLM-7B.

\section{Related Work}
\label{sec:related}

\textbf{Token compression} reduces visual sequence length within the vision encoder, at the projector between the encoder and the LLM, or within the LLM~\citep{wu2026data}. Methods select tokens by visual saliency~\citep{shang2025llava} or language-conditioned relevance~\citep{xing2024pyramiddrop}, merge similar tokens~\citep{bolya2023tome}, or combine selection with aggregation~\citep{yang2025visionzip}. For multi-view 3D inputs, geometric correspondence guides cross-view redundancy reduction, while spatial diversity promotes scene coverage. DTC~\citep{huang2025dtc} and ToSA~\citep{huang2025tosa} guide merging through voxel constraints and depth-derived spatial tokens, respectively. SeGPruner~\citep{li2026segpruner} and Geo3DPruner~\citep{li2026geometry} select geometrically diverse tokens, while Prune2Drive~\citep{xiong2025prune2drive} combines diversity-aware sampling with view-adaptive budgets. Within the LLM, Fast3D~\citep{huang2025fast3d} assigns sample-adaptive pruning budgets, and AdaToken-3D~\citep{zhang2025adatoken3d} schedules layer-wise token retention using spatial information flow. HiSC~\citep{qu2026hisc} combines spatial graph-based merging before the LLM with hierarchical cluster-based pruning within it. Our GeoSemZip combines salient token selection with spatial coverage and aggregates unselected voxel features into contextual anchors to retain additional scene information.


\textbf{Operation compression} reduces computation along LLM width and depth~\citep{wu2026data, hu2026wide}. In image-based MLLMs, RedundancyLens~\citep{li2025redundancylens} and YOPO~\citep{zhang2024treat} exploit redundancy within attention and FFNs, with YOPO also pruning deep-layer visual computation. MWOP~\citep{wang2026mwop} further reduces computation within layers by pruning modality-interaction paths in attention heads and selecting FFN channels separately for visual and textual tokens while preserving the token sequence. Along depth, Skip-Vision~\citep{zeng2025skip} skips FFN updates for selected visual tokens, whereas ShortV~\citep{yuan2025shortv} freezes visual token updates in low-contribution layers. HiDrop~\citep{wu2026hidrop} combines late injection and early exit with progressive token pruning. ViCA~\citep{liu2026vica} bypasses visual token updates and provides text access to static visual features through sparse cross-attention. Our GroupRoute extends late entry and early exit to 3D VLMs with recoverable group-wise skipping. Query-conditioned token groups bypass complete LLM layers with their states preserved and are reassessed at subsequent anchor layers for reactivation.

\section{Understanding Layer-Wise Visual Token Dynamics in 3D VLMs}
\label{sec:layer_dynamics}

Visual tokens play different roles across LLM layers, with vision--language interaction concentrated in the middle layers~\citep{wu2026hidrop}. This motivates late visual entry and early visual exit, which restrict visual computation to a middle-layer window (Section~\ref{sec:llm}). Within this window, token rankings may still change as vision--language interaction proceeds, making it unclear whether a single-layer ranking provides a reliable basis for permanent token pruning. We therefore examine the cross-layer stability of attention-based visual token rankings, focusing on their consistency across nearby layers and whether initially low-ranked tokens become highly ranked at later layers.

\begin{wrapfigure}{r}{0.5\textwidth}
    \vspace{-15pt}
    \centering
    \includegraphics[width=\linewidth]{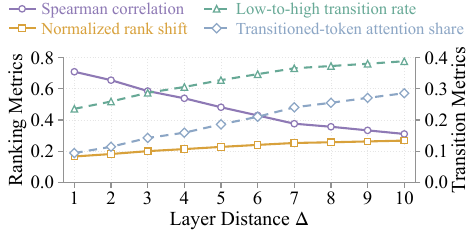}
    \caption{\small \textbf{Cross-layer stability of visual token rankings.} Ranking stability decreases with layer distance, while initially low-ranked tokens increasingly enter the top half and account for a growing share of its attention.}
    \label{fig:token_rank_dynamics}
    \vspace{-1em}
\end{wrapfigure}

\paragraph{Tracking Cross-layer Ranking Stability.}
We conduct our analysis on ScanQA~\citep{azuma2022scanqa} using LLaVA-OneVision, focusing on layers 8--23 within the middle-layer visual computation window. At layer $l$, we assign each visual token $i$ an attention score using the attention from the last prompt token $q$ before answer generation, averaged over all $H$ attention heads:
\begingroup
\setlength{\abovedisplayskip}{3pt}
\setlength{\belowdisplayskip}{3pt}
\setlength{\abovedisplayshortskip}{3pt}
\setlength{\belowdisplayshortskip}{3pt}
\begin{equation*}
    s_i^{(l)}
    =
    \frac{1}{H}
    \sum_{h=1}^{H}
    A_{h,q,i}^{(l)},
\end{equation*}
\endgroup
where $A_{h,q,i}^{(l)}$ is the attention weight from $q$ to $i$ in head $h$. We rank tokens by $s_i^{(l)}$ and split them into top and bottom halves. For each layer pair $(l,l+\Delta)$, we compute Spearman correlation and normalized rank shift, alongside the fraction of later top-half tokens originating from the earlier bottom half and their attention share within the later top half. We average these metrics over all valid layer pairs at each distance $\Delta=1,\ldots,10$. Figure~\ref{fig:token_rank_dynamics} shows two complementary patterns: rankings are more stable across nearby layers, while initially low-ranked tokens increasingly enter the top half and contribute a growing share of its attention at larger layer distances. See Appendix~\ref{app:ranking_heatmaps} for full layer-pair heatmaps.

\paragraph{Ranking Stability over Layer Distance.}
Figure~\ref{fig:token_rank_dynamics} shows ranking stability through two metrics on the left y-axis: Spearman correlation measures agreement between complete token rankings, while normalized rank shift measures the mean absolute rank change divided by the maximum rank span. Nearby layers exhibit higher rank correlation and smaller rank shifts, whereas agreement generally weakens as layer distance increases. These results indicate that token rankings retain local consistency but cannot be assumed to remain stable throughout the visual computation window. This local consistency motivates reusing a token partition over a short layer interval to reduce the overhead of repeated scoring and routing.

\paragraph{Low-to-High Rank Transitions.}
The two metrics on the right y-axis of Figure~\ref{fig:token_rank_dynamics} quantify how initially low-ranked tokens enter the top half at later layers. The transition rate measures their fraction of the later top half, while the attention share measures their contribution to that group's total attention. As layer distance increases, a growing portion of the later top half originates from initially low-ranked tokens. Crucially, these tokens also account for a growing, non-negligible share of the attention assigned to that group. A low rank at one layer therefore does not guarantee a low rank at subsequent layers. This motivates treating low-ranked tokens as \emph{temporarily inactive at the current stage}, rather than \emph{permanently redundant in all subsequent layers}, and retaining them for later reassessment.

\paragraph{Implications for Group-wise Computation.}
Together, these observations motivate reusing token partitions over short layer intervals while preserving the opportunity for reassessment. Local ranking stability motivates temporary reuse of a partition, while low-to-high transitions motivate retaining low-ranked tokens for later recovery and regrouping. This leads to the group-wise skip mechanism in Section~\ref{sec:llm}, where low-ranked tokens temporarily skip LLM blocks while preserving their hidden states. At the next anchor layer, all visual tokens, including those previously skipped, undergo full LLM-block computation and are rescored to determine their groups for the following interval.
\section{Method}
\label{sec:method}

\begin{figure*}[t]
    \centering
    \includegraphics[width=\textwidth]{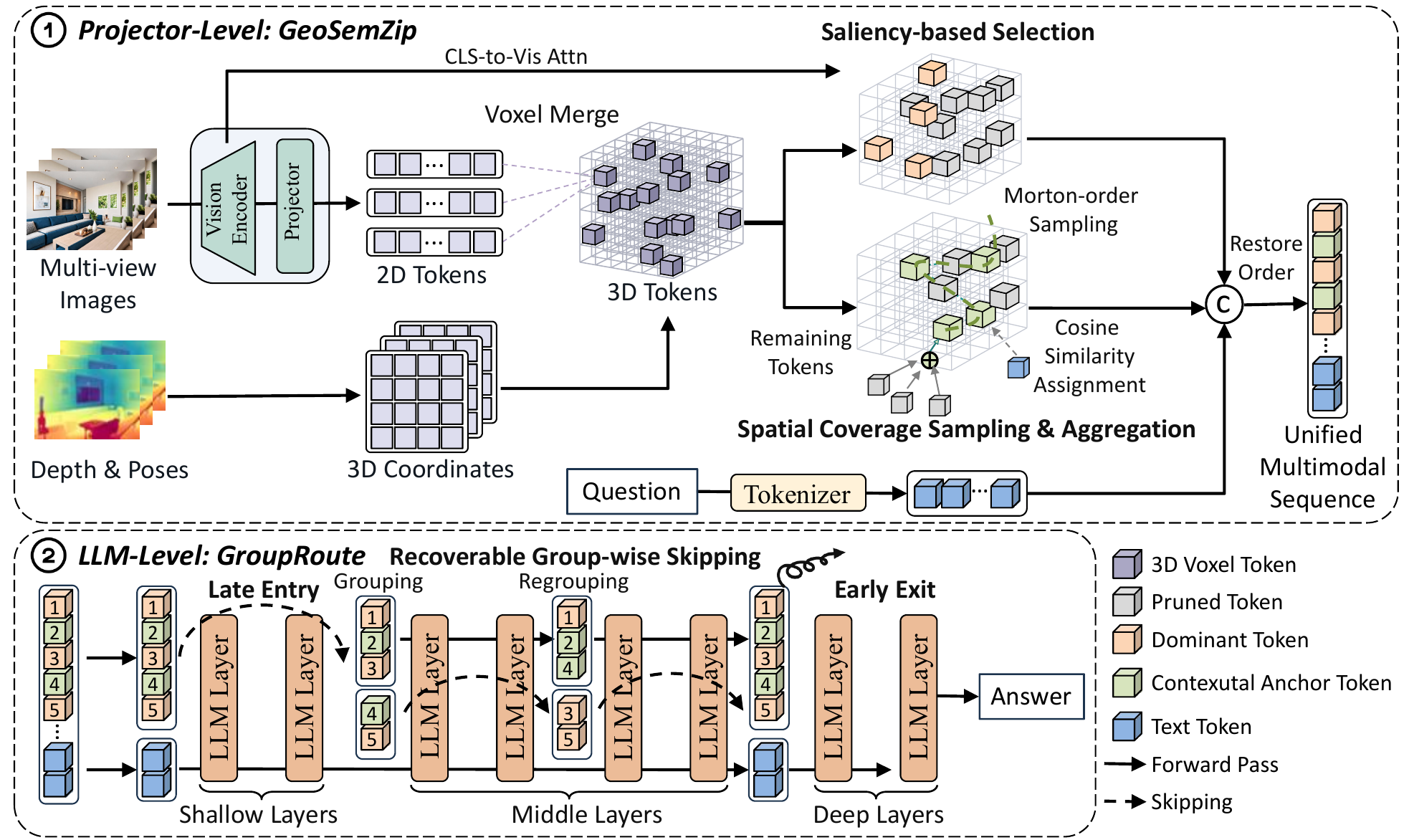}
    \caption{\small
    \textbf{Overview of GeoGR}.
    \textbf{(1) Projector-level Compression:}
    GeoSemZip consolidates cross-view observations into 3D voxel tokens
    using depth and camera poses. Saliency-based selection preserves
    dominant tokens, while Morton-order sampling selects contextual
    anchors that aggregate unselected voxel features by cosine similarity.
    The compressed visual tokens are combined with text tokens to form
    the LLM input.
    \textbf{(2) LLM-level Compression:}
    GroupRoute confines visual computation to an intermediate-layer
    window through Late Entry and Early Exit. Within this window,
    Recoverable Group-wise Skipping routes query-conditioned token groups
    through or around complete LLM layers. Skipped tokens retain their
    hidden states and can resume computation after regrouping at
    subsequent anchor layers. Text tokens remain active throughout the LLM.
    }
    \label{fig:overview}
\end{figure*}

As illustrated in Figure~\ref{fig:overview}, \emph{GeoGR} combines projector-level and LLM-level compression for multi-view 3D VLMs. \emph{GeoSemZip} consolidates cross-view observations into compact tokens preserving salient content and spatial coverage. \emph{GroupRoute} combines \emph{Late Entry and Early Exit} to define a visual computation window with \emph{Recoverable Group-wise Skipping} to selectively allocate computation according to query relevance while preserving skipped token states for later reactivation. We first present the distillation dataset and post-training protocol, followed by the two compression stages.

\subsection{Post-Training Protocol for 3D VLM Compression}
\label{sec:pipeline}

\begin{wraptable}{r}{0.47\textwidth}
    \vspace{-10pt}
    \centering
    \footnotesize
    \setlength{\tabcolsep}{5pt}
    \renewcommand{\arraystretch}{1.0}
    \caption{\small Our post-training dataset distilled by selecting 30\% of samples from each source using PRISM (top), and performance retained relative to full-data post-training (bottom).}
    \label{tab:posttrain_data}
    \begin{tabular}{@{}lrr@{}}
        \toprule
        \textbf{Source} & \textbf{Full Set} & \textbf{Selected} \\
        \midrule
        ScanRefer    & 36,665  & 11,000 \\
        Multi3DRefer & 43,838  & 13,151 \\
        Scan2Cap     & 36,665  & 11,000 \\
        ScanQA       & 26,515  & 7,955  \\
        SQA3D        & 79,445  & 23,834 \\
        \midrule
        \textbf{Total}        & 223,128 & 66,940 \\
        \midrule
        \textbf{3D VLMs} &
        \multicolumn{2}{c}{\textbf{Retention (\%)}} \\
        \midrule
        Video-3D LLM-7B    & \multicolumn{2}{c}{98.72} \\
        LLaVA-OneVision-7B & \multicolumn{2}{c}{95.69} \\
        \bottomrule
    \end{tabular}
    \vspace{-5pt}
\end{wraptable}

We collect 223K training examples from ScanRefer~\citep{chen2020scanrefer}, Multi3DRefer~\citep{zhang2023multi3drefer}, Scan2Cap~\citep{chen2021scan2cap}, ScanQA~\citep{azuma2022scanqa}, and SQA3D~\citep{ma2023sqa3d}. To reduce post-training cost, we construct a distilled dataset using PRISM~\citep{bi2026prism}, a training-free data selection method that estimates redundancy through the average correlation between mean-centered intrinsic visual representations. We retain the least redundant 30\% of samples from each source, yielding a 67K-example post-training dataset, as summarized in Table~\ref{tab:posttrain_data}. To assess the subset's effectiveness, we post-train dense models without any compression enabled. Relative to their counterparts post-trained on the full dataset, Video-3D LLM-7B and LLaVA-OneVision-7B retain 98.72\% and 95.69\% of performance, respectively. Detailed results for dense models post-trained on the full and distilled datasets are provided in Appendix~\ref{app:posttrain_results}. This dataset supports a unified compression-aware post-training protocol for adapting models to compression.

\subsection{GeoSemZip: Spatial--Semantic Token Compression}
\label{sec:projector_method}

\emph{GeoSemZip} constructs a compact scene representation preserving salient content and spatial coverage under a fixed token budget. It consolidates cross-view observations into 3D voxel tokens, then divides the budget between salient dominant tokens and spatially distributed contextual anchors. Remaining voxel features are aggregated into the contextual anchors based on feature similarity.

\paragraph{Cross-view Voxel Consolidation.}
Given $T$ calibrated views with aligned depth, let $\{\mathbf{x}_i\}_{i=1}^{N}$ denote the projected visual tokens, where $\mathbf{x}_i\in\mathbb{R}^{d}$ and $N=THW$ for an $H\times W$ grid per view. Back-projection using depth, intrinsics, and camera-to-world poses assigns each token a world coordinate $\mathbf{p}_i$. Saliency $a_i$ is computed from received attention in the last executed vision-encoder layer, averaged over heads and query positions and pooled to the projected grid. Voxelization with width $s$ gives $\mathbf{g}_i=\operatorname{round}(\mathbf{p}_i/s)$. Let $\mathcal{V}$ be the occupied voxel set and $\mathcal{G}_v=\{i:\mathbf{g}_i=v\}$ the token indices in voxel $v$. We aggregate each voxel's features by averaging and saliency by max pooling:
\begingroup
\small
\setlength{\abovedisplayskip}{3pt}
\setlength{\belowdisplayskip}{3pt}
\setlength{\abovedisplayshortskip}{3pt}
\setlength{\belowdisplayshortskip}{3pt}
\begin{equation*}
\mathbf{z}_v =
\frac{1}{|\mathcal{G}_v|}
\sum_{i\in\mathcal{G}_v}\mathbf{x}_i,
\qquad
b_v = \max_{i\in\mathcal{G}_v} a_i
\end{equation*}
\endgroup
Feature averaging consolidates spatially co-located observations across views, while max pooling preserves the strongest saliency response for subsequent token selection.

\paragraph{Saliency- and Coverage-aware Selection.}
Given $K=|\mathcal{V}|$ occupied voxels and a target keep ratio $r$, we set the token budget to $M=\min(K,\lceil rN\rceil)$. We allocate a fraction $\rho$ to the highest-saliency dominant tokens and the remainder to contextual anchors: 
\begingroup
\small
\setlength{\abovedisplayskip}{3pt}
\setlength{\belowdisplayskip}{3pt}
\setlength{\abovedisplayshortskip}{3pt}
\setlength{\belowdisplayshortskip}{3pt}
\begin{equation*}
M_d=\operatorname{clip}\!\left(
\operatorname{round}(\rho M),1,M-1\right),
\quad
M_c=M-M_d,
\quad
\mathcal{D}=\operatorname{TopK}_{v\in\mathcal{V}}(b_v,M_d)
\end{equation*}
\endgroup
For spatial coverage, we select contextual anchors from the remaining voxels $\mathcal{R}=\mathcal{V}\setminus\mathcal{D}$. Their integer coordinates are shifted to be nonnegative and
sorted by 3D Morton code, producing a sequence that approximately preserves spatial locality. We sample $M_c$ anchors at a fixed stride $\lfloor|\mathcal{R}|/M_c\rfloor$ to form $\mathcal{C}$. This selection is independent of saliency scores, encouraging coverage of less salient regions.

\paragraph{Residual Merging and Serialization.}
To retain information from voxels selected by neither branch, we aggregate their features into contextual anchors. Each unselected voxel is assigned to the anchor whose feature has the highest cosine similarity to its own. Each anchor then adds the mean feature of its assigned voxels as a residual, incorporating unselected content without increasing the token budget. Anchors with no assigned voxels and all dominant tokens remain unchanged. For serialization, each retained voxel inherits the index of its earliest contributing input token. We sort the retained tokens by these indices, producing an ordered sequence of $M$ visual tokens for the LLM.

\subsection{GroupRoute: Late Entry, Early Exit, and Recoverable Group-wise Skipping}
\label{sec:llm}

After \emph{GeoSemZip}, \emph{GroupRoute} uses \emph{Late Entry} and \emph{Early Exit} to define a visual computation window, within which \emph{Recoverable Group-wise Skipping (RGS)} selectively updates tokens according to query relevance while preserving skipped token states for later reactivation. \emph{Budget-constrained Anchor and Regrouping Search (BARS)} selects the regrouping anchors under a computation budget.

\begin{figure*}[t]
    \centering
    \includegraphics[width=\textwidth]{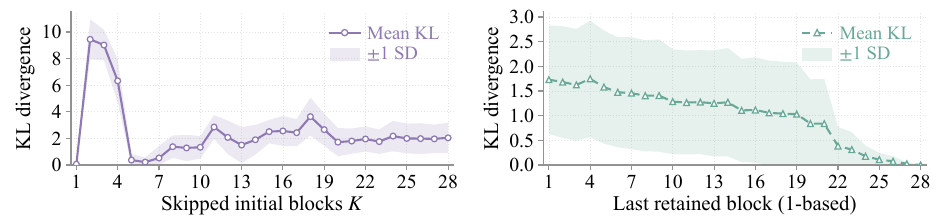}
    \caption{\small Visual-window diagnosis for LLaVA-OneVision on 256 ScanQA examples: Late Entry (left) and Early Exit (right). Lines show mean next-token KL divergence relative to full visual computation; shading indicates $\pm1$ standard deviation, with the lower bound clipped at zero.}
    \label{fig:visual_window_kl}
\end{figure*}

\paragraph{Late Entry and Early Exit.}
Following HiDrop~\citep{wu2026hidrop}, we adapt late entry and early exit to multi-view 3D VLMs. Text tokens pass through all LLM layers, while visual tokens participate only within a window $[\ell_{\mathrm{in}},\ell_{\mathrm{out}}]$. They enter at layer $\ell_{\mathrm{in}}$ with their original full-prompt position IDs and exit after layer $\ell_{\mathrm{out}}$. Both boundaries use one-based indexing. To identify candidate boundaries, we independently vary the number of initial blocks skipped and the last block retaining visual tokens, measuring next-token KL divergence relative to full visual computation. As shown in Figure~\ref{fig:visual_window_kl}, skipping the first $K=5$--$7$ blocks yields the final consecutive low-mean-KL region before divergence increases. For early exit, KL drops sharply between retained blocks 21 and 22 and remains low from block 23 onward. These pre-adaptation measurements define candidate ranges, which we refine through compression-aware post-training. The resulting window for LLaVA-OneVision is $[8,23]$, as evaluated in the window ablation study.

\paragraph{Recoverable Group-wise Skipping.}
Within the visual-computation window, we partition visual tokens at designated anchor layers according to their relevance to the query. The query is the last prompt token for question answering and captioning, or the grounding token for localization. At anchor $a$, we reuse the layer's query and key projections and positional encoding to compute head-averaged attention scores from its input representations:
\begingroup
\small
\setlength{\abovedisplayskip}{3pt}
\setlength{\belowdisplayskip}{3pt}
\setlength{\abovedisplayshortskip}{3pt}
\setlength{\belowdisplayshortskip}{3pt}
\begin{equation*}
s_i^{(a)}
=
\frac{1}{H_q}\sum_{h=1}^{H_q}
\left[
\operatorname{softmax}\!\left(
\frac{\mathbf{q}_{a,h}\mathbf{K}_{a,h}^{\top}}
{\sqrt{d_h}}
\right)
\right]_i
\end{equation*}
\endgroup
Here $H_q$ is the number of query heads, $d_h$ is the head dimension, and the softmax covers all causally visible prompt tokens. The highest-scoring $\lceil\kappa N_{\mathrm{vis}}\rceil$ visual tokens form the high-priority group $\mathcal{H}_a$, and the remaining tokens form the low-priority group $\mathcal{L}_a$, where $N_{\mathrm{vis}}$ is the number of visual tokens retained by GeoSemZip. All visual tokens execute the anchor layer; the partition takes effect only in subsequent layers. Until the next anchor, text tokens and $\mathcal{H}_a$ execute complete LLM layers with their original position IDs, while $\mathcal{L}_a$ bypasses both attention and FFN computation and contributes no keys or values. The skipped tokens retain their hidden states and positions. At the next anchor, all visual tokens are rescored and updated, allowing previously inactive tokens to rejoin the high-priority group. The final partition remains in effect until early exit.

\paragraph{Budget-constrained Anchor Selection.}
BARS selects anchor positions under a computation budget using layer-wise attention scores normalized over retained visual tokens. For a partition formed at anchor $a$ and reused at layer $\ell$, we quantify three risks: low-priority attention mass, migration from the low-priority to high-priority group, and mean normalized rank displacement. Risks are weighted by the remaining visual-computation depth, averaged within benchmarks, and then equally across benchmarks. For each feasible anchor count $m$, dynamic programming finds the minimum-risk schedule $\mathcal A_m$ under a normalized visual token--layer budget and minimum anchor gap. Starting with $\mathcal A_1$, BARS considers increasing anchor counts, accepting a candidate only if it reduces both aggregate risk and each calibration subset's risk by at least $\tau=5\%$ relative to the current schedule. With $\kappa=0.5$, BARS selects anchors $[9,13]$ on both 3D VLMs. The schedule is fixed at inference, while token groups remain query-dependent. Appendix~\ref{app:bars} provides the risk definitions and procedure.

\section{Experiments}
\label{sec:experiments}

\subsection{Experimental setup}
\label{sec:experimental_setup}

\paragraph{Models.}
We evaluate GeoGR on two representative multi-view 3D VLMs: LLaVA-OneVision-7B~\citep{li2024llavaov}, which independently encodes views without explicit 3D representations, and Video-3D LLM-7B~\citep{zheng2024video3dllm}, which augments view features with depth- and pose-derived 3D positional embeddings in a shared scene coordinate frame. Both models take 32 uniformly sampled RGB views per scene. Architectural details are provided in Appendix~\ref{app:3dvlm}.

\paragraph{Benchmarks.}
We evaluate GeoGR on five ScanNet-based benchmarks covering visual grounding (ScanRefer~\citep{chen2020scanrefer} and Multi3DRefer~\citep{zhang2023multi3drefer}), dense captioning (Scan2Cap~\citep{chen2021scan2cap}), and 3D question answering (ScanQA~\citep{azuma2022scanqa} and SQA3D~\citep{ma2023sqa3d}). Video-3D LLM is evaluated on all five, while LLaVA-OneVision is evaluated only on the two QA benchmarks because it does not support the required 3D localization outputs. See Appendix~\ref{app:bench} for details.

\paragraph{Efficiency Evaluation.}
Since FLOP reductions do not necessarily translate into proportional inference speedups~\citep{hu2026beyond}, we evaluate both computational cost and measured latency. To estimate visual FLOPs, let $n_\ell$ be the number of active visual tokens at layer $\ell$, $d$ the hidden dimension, $m$ the gated FFN intermediate dimension, and $r=H_{\mathrm{KV}}/H_{\mathrm{Q}}$ the key/value-to-query head ratio. The Q/K/V/O projection widths are $d$, $rd$, $rd$, and $d$. Counting multiply-adds as two FLOPs  and using dense $n_\ell\times n_\ell$ visual attention, we obtain
\begingroup
\small
\setlength{\abovedisplayskip}{3pt}
\setlength{\belowdisplayskip}{3pt}
\setlength{\abovedisplayshortskip}{3pt}
\setlength{\belowdisplayshortskip}{3pt}
\begin{equation*}
  \operatorname{FLOPs}_{\mathrm{vis}}
  \approx 2\sum_{\ell=1}^{L}
  \left[(2+2r)n_\ell d^2+2n_\ell^2d+3n_\ell dm\right]
\end{equation*}
\endgroup
Both 3D VLMs use Qwen2-7B with $H_{\mathrm{Q}}=28$ and
$H_{\mathrm{KV}}=4$, giving $r=1/7$.

\paragraph{Implementation Details.}
For LLaVA-OneVision, we pool the projector features of each view to a $14\times14$ grid, yielding $6{,}272$ visual tokens per scene before compression. Unless otherwise stated, GeoSemZip uses voxel size $s=0.1$, target keep ratio $r=0.30$, and dominant-token ratio $\rho=0.85$, with max pooling for voxel saliency scores. The visual computation windows are $[8,23]$ for LLaVA-OneVision and $[8,24]$ for Video-3D LLM, using one-based, inclusive layer indices.  BARS selects anchor layers 9 and 13 for both models, with a keep ratio $\kappa=0.5$ at each anchor. Compression-aware post-training has two stages. The first enables GeoSemZip and trains the projector and full LLM with the vision encoder frozen. The second enables Late Entry and Early Exit and trains only the full LLM, keeping the vision encoder and projector frozen. All evaluations use greedy decoding (temperature 0, one beam) with at most 512 new tokens. Complete configurations are provided in Appendix~\ref{sec:training_protocol}.


\begin{table*}[t]
    \caption{\small
    \textbf{Performance on LLaVA-OneVision-7B and Video-3D LLM-7B at approximately 10\% retained LLM-side visual FLOPs.} LLaVA-OneVision is evaluated only on ScanQA and SQA3D, as it does not support the 3D localization outputs required by the other benchmarks. ``Avg.'' measures performance retention relative to the dense model. ``Tokens'' denotes the average number of active visual tokens per LLM layer. Unshaded baseline rows apply compression without recovery, while light-red rows use compression-aware post-training. Light-blue rows denote GeoGR. $^\dagger$ marks baselines originally proposed without recovery and evaluated with it here. $^*$ marks compression baselines originally designed for 3D inputs, while the remaining baselines were designed for 2D inputs. \textbf{Bold} and \underline{underlined} values indicate the \textbf{best} and \underline{second-best} results among compressed methods, respectively. GeoGR achieves the highest average performance retention on both 3D VLMs at comparable computational budgets.
    }
  \centering
  \scriptsize
  \setlength{\tabcolsep}{2.0pt}
  \resizebox{\textwidth}{!}{%
  \begin{tabular}{@{}l|cc|ccccccccc|c@{}}
    \toprule
    \multirow{2}{*}{\textbf{Method}}
      & \multicolumn{2}{c|}{\textbf{Budget}}
      & \multicolumn{2}{c}{\textbf{ScanQA}}
      & \textbf{SQA3D}
      & \multicolumn{2}{c}{\textbf{Scan2Cap}}
      & \multicolumn{2}{c}{\textbf{ScanRefer}}
      & \multicolumn{2}{c|}{\textbf{Multi3DRefer}}
      & \multirow{2}{*}{\textbf{Avg.}} \\
    \cmidrule(lr){2-3} \cmidrule(lr){4-5} \cmidrule(lr){6-6} \cmidrule(lr){7-8} \cmidrule(lr){9-10} \cmidrule(lr){11-12}
      & \textbf{Tokens} & \textbf{TFLOPs}
      & \textbf{CIDEr} & \textbf{EM} &  \textbf{EM}
      & \textbf{B-4@.5} & \textbf{C@.5}
      & \textbf{Acc@.25} & \textbf{Acc@.5}
      & \textbf{F1@.25} & \textbf{F1@.5} & \textbf{(\%)} \\
    \midrule
    \rowcolor{gray!20}
    \multicolumn{13}{c}{\textbf{LLaVA-OneVision-7B}} \\
    \midrule
    Dense & 6272 & 97.64 & 94.84 & 28.09 & 57.06 & & & & & & & 100.00 \\
    \midrule
    VisionZip~\citep{yang2025visionzip}     & 640 & 8.52 & 78.54 & 23.14 & 51.41 & - & - & - & - & - & - & 85.10 \\
    VisPruner~\citep{zhang2025beyond}    & 640 & 8.52 & 80.37 & 23.81 & 52.09 & - & - & - & - & - & - & 86.93 \\
    SeGPruner$^{*}$~\citep{li2026segpruner}       & 640 &  8.52 & 85.15 & 25.43 & 53.88 & - & - & - & - & - & - & 91.59 \\
    VTC$^{*}$~\citep{huang2025dtc}                & 735 & 9.81 & 85.64 & 25.33 & 54.36 & - & - & - & - & - & - & 91.91 \\
    DTC$^{*}$~\citep{huang2025dtc}                & 702 & 9.36 & \second{87.22} & \second{25.81} & 54.78 & - & - & - & - & - & - & \second{93.30} \\
    \rowcolor{red!8}
    VisionZip$^{\dagger}$~\citep{yang2025visionzip} & 640 & 8.52 & 81.36 & 23.59 & 54.33 & - & - & - & - & - & - & 88.34 \\
    \rowcolor{red!8}
    VisPruner$^{\dagger}$~\citep{zhang2025beyond}& 640 & 8.52 & 84.67 & 25.09 & \second{56.01}& - & - & - & - & - & - & 92.26 \\
    \rowcolor{red!8}
    SeGPruner$^{*, \dagger}$~\citep{li2026segpruner}   & 640 & 8.52 & 84.52 & 24.86 & 55.61 & - & - & - & - & - & - & 91.69 \\
    \rowcolor{red!8}
    DTC$^{*, \dagger}$~\citep{huang2025dtc}            & 702 & 9.36 & 86.06 & 25.50 & 54.96 & - & - & - & - & - & - & 92.61 \\
    \rowcolor{oursblue}
    GeoGR (ours) & 638 & 8.65 & \best{88.25} & \best{26.20} & \best{56.61} & - & - & - & - & - & - & \best{95.18} \\
    
    \midrule
    \rowcolor{gray!20}
    \multicolumn{13}{c}{\textbf{Video-3D LLM-7B}} \\
    \midrule
    Dense & 6272 & 97.64 & 98.19 & 27.74 & 57.43 & 40.71 & 80.02 & 57.79 & 51.48 & 56.95 & 51.75 & 100.00 \\
    \midrule
       VisionZip~\citep{yang2025visionzip} & 704 & 9.39 & 89.60 & 25.11 & 55.30 & 37.00 & 59.56 & 55.72 & 49.78 & 53.40 & 48.67 & 91.59 \\
    VisPruner~\citep{zhang2025beyond} & 704 & 9.39 & 88.19 & 24.64 & 54.67 & 36.71 & 58.55 & 55.63 & 49.76 & 53.01 & 48.39 & 90.74 \\
     SeGPruner$^{*}$~\citep{li2026segpruner} & 640 & 8.52 & 88.91 & 25.07 & 55.10 & 36.04 & 57.99 & 56.11 & 49.96 & 53.99 & 49.11 & 91.30 \\
    
    VTC$^{*}$~\citep{huang2025dtc} & 708 & 9.44 & 89.14 & 25.26 & 54.16 & 36.36 & 57.07 & 55.92 & 49.65 & 52.67 & 47.89 & 90.56 \\

    DTC$^{*}$~\citep{huang2025dtc} & 704 & 9.39 & 87.06 & 24.11 & 54.11 & 36.60 & 58.28 & 55.50 & 49.45 & 51.57 & 47.00 & 89.55 \\

    \rowcolor{red!8}
    VisionZip$^{\dagger}$~\citep{yang2025visionzip} & 704 & 9.39 & 93.30 & \second{26.42} & \best{56.72} & 38.78 & 70.42 & \second{56.69} & 50.37 & 55.76 & \second{50.71} & \second{96.01} \\
    \rowcolor{red!8}
    VisPruner$^{\dagger}$~\citep{zhang2025beyond} & 704 & 9.39 & 92.60 & 26.31 & \second{56.49} & \second{39.23} & 70.15 & 56.52 & 50.21 & \second{55.86} & 50.66 & 95.87 \\
    \rowcolor{red!8}
    SeGPruner$^{*, \dagger}$~\citep{li2026segpruner} & 640 & 8.52 & \second{94.23} & \best{26.46} & 55.90 & 38.77 & 70.66 & 56.33 & 50.11 & 55.62 & 50.41 & 95.79 \\ 
    \rowcolor{red!8}
    DTC$^{*, \dagger}$~\citep{huang2025dtc} & 704 & 9.39 & 91.28 & 25.67 & 56.29 & 39.13 & \second{70.98} & 56.65 & \second{50.39} & 55.41 & 50.29 & 95.41 \\
    \rowcolor{oursblue}
    GeoGR (ours) & 671 & 9.09 & \best{94.85} & 26.22 & 56.46 & \best{39.45} & \best{73.27} & \best{56.87} & \best{50.69} & \best{55.89} & \best{50.84} & \best{96.80} \\
    \bottomrule
  \end{tabular}
  }
  \label{tab:main}
\end{table*}

\subsection{Main results}
\label{sec:main_results}

\paragraph{Comparison with State-of-the-art Methods.}
Table~\ref{tab:main} compares GeoGR with representative 2D and 3D compression methods at approximately 10\% retained LLM-side visual FLOPs. On LLaVA-OneVision, GeoGR retains 95.18\% of the dense model's performance, exceeding the strongest baseline, DTC, by 1.88 percentage points while requiring fewer FLOPs (8.65 vs.\ 9.36 TFLOPs). On Video-3D LLM, we evaluate five benchmarks covering QA, visual grounding, and dense captioning. GeoGR achieves the highest average performance retention of 96.80\% at 9.09 TFLOPs. Compared with the strongest baseline, VisionZip$^{\dagger}$, it improves retention by 0.79 percentage points while reducing FLOPs by 3.2\%. The results also suggest that compression effectiveness depends on how spatial information is incorporated. After compression-aware post-training, the 2D-native methods VisionZip and VisPruner retain 96.01\% and 95.87\% of performance on Video-3D LLM, compared with 88.34\% and 92.26\% on LLaVA-OneVision. Although these averages cover different benchmark sets, a plausible explanation is that Video-3D LLM embeds depth- and pose-derived 3D positional information into its visual representations, potentially benefiting even compressors designed for 2D inputs. LLaVA-OneVision lacks this explicit spatial encoding, but 3D compression methods can still use geometric information to guide token selection and aggregation. Such external guidance may be particularly valuable for preserving cross-view evidence on LLaVA-OneVision, consistent with the stronger results of 3D-native baselines on this model. GeoGR achieves strong performance retention on both 3D VLMs, supporting its applicability to models with different scene representations.

\begin{wraptable}{r}{0.52\textwidth}
  \centering
  \vspace{-15pt}
  \caption{\small
  \textbf{Efficiency comparison on two 3D VLMs.}
  Parentheses indicate speedups relative to the dense model.
  ``Projector'' covers all operations between the vision encoder
  and the LLM, including token compression.
  Encoder latency is similar across methods and is omitted
  from the breakdown but included in TTFT.
  GeoGR offers a favorable trade-off between performance retention
  and inference efficiency on both 3D VLMs.
  }
  \label{tab:efficiency}
  \scriptsize
  \setlength{\tabcolsep}{1.5pt}
  \resizebox{\linewidth}{!}{%
  \begin{tabular}{c|c|ccc|c}
    \toprule
    \multirow{2}{*}{\textbf{Method}}
    & \multirow{2}{*}{\makecell{\textbf{KV Cache}\\\textbf{(MB)}}}
    & \multicolumn{3}{c|}{\textbf{Latency (ms)}}
    & \multirow{2}{*}{\makecell{\textbf{Avg.}\\\textbf{(\%)}}} \\
    & &
    \textbf{Projector}
    & \textbf{LLM Prefill}
    & \textbf{TTFT}
    & \\
    \midrule
    \rowcolor{gray!20}
    \multicolumn{6}{c}{\textbf{LLaVA-OneVision-7B}} \\
    \midrule
    Dense
    & 359.7 & 53.3
    & 1021.4 ($1.0\times$)
    & 1426.7 ($1.0\times$)
    & 100.0 \\
    SeGPruner$^{\dagger}$
    & 36.7 & 100.8
    & 103.0 ($9.9\times$)
    & 548.5 ($2.6\times$)
    & 91.7 \\
    DTC$^{\dagger}$
    & 40.3 & 2025.0
    & 147.2 ($6.9\times$)
    & 2522.2 ($0.6\times$)
    & 92.6 \\
    \rowcolor{oursblue}
    \textbf{GeoGR}
    & 36.6 & 99.0
    & 151.2 ($6.8\times$)
    & 605.0 ($2.4\times$)
    & 95.2 \\
    \midrule
    \rowcolor{gray!20}
    \multicolumn{6}{c}{\textbf{Video-3D LLM-7B}} \\
    \midrule
    Dense
    & 359.7 & 65.2
    & 874.1 ($1.0\times$)
    & 1257.8 ($1.0\times$)
    & 100.0 \\
    SeGPruner$^{\dagger}$
    & 36.7 & 76.4
    & 86.4 ($10.1\times$)
    & 479.7 ($2.6\times$)
    & 95.8 \\
    DTC$^{\dagger}$
    & 40.3 & 2301.1
    & 132.4 ($6.6\times$)
    & 2750.6 ($0.5\times$)
    & 95.4 \\
    \rowcolor{oursblue}
    \textbf{GeoGR}
    & 38.5 & 75.6
    & 143.4 ($6.1\times$)
    & 537.4 ($2.3\times$)
    & 96.8 \\
    \bottomrule
  \end{tabular}%
  }
  \vspace{-10pt}
\end{wraptable}

\paragraph{Efficiency Analysis.}
As shown in Table~\ref{tab:efficiency}, GeoGR speeds up LLM prefill by $6.8\times$ on LLaVA-OneVision and $6.1\times$ on Video-3D LLM, with TTFT speedups of $2.4\times$ and $2.3\times$. It reduces the visual token KV cache from 359.7\,MB to 36.6 and 38.5\,MB, respectively. Practical acceleration depends on costs across the inference pipeline, as illustrated by the stage-wise bottleneck analysis in ViCoStream~\citep{tan2026vicostream}. In our setting, compression overhead is critical to realizing these gains. DTC spends over 2000\,ms in the projector stage, outweighing its prefill savings, whereas GeoGR takes only 99.0 and 75.6\,ms. These efficiency gains should also be considered alongside the performance retained after compression. SeGPruner is faster, but GeoGR retains more performance, especially when a 2D VLM is used for 3D scene understanding. On LLaVA-OneVision, GeoGR retains 95.2\% of performance compared with 91.7\% for SeGPruner, a gain of 3.5 percentage points for an additional 56.5\,ms of TTFT. These results show a favorable trade-off between performance retention and inference efficiency.

\subsection{Ablation studies}
\label{sec:ablation}

We conduct ablation studies on LLaVA-OneVision-7B to examine the components of GeoSemZip, the grouping signal, and recoverable skipping. Avg. reports performance retention relative to the dense post-trained model (100\%). Additional ablations on voxel size, saliency aggregation, the dominant-token ratio, and the visual computation window are provided in Appendix~\ref{app:exp}.

\begin{wraptable}{r}{0.47\textwidth}
    \centering
    \vspace{-15pt}
    \caption{ \small
    \textbf{Component ablation of GeoSemZip.} All variants retain 30\% of the visual tokens and are post-trained independently from the same initialization. The semantic-only variant omits cross-view voxel consolidation. Residual merging requires contextual anchors from the spatial coverage branch.
    } 
    \label{tab:projector_component_ablation}
    \setlength{\tabcolsep}{2pt}
    \resizebox{\linewidth}{!}{
    \begin{tabular}{ccc|ccc|c}
        \toprule
        \textbf{Semantic} & \textbf{Spatial} & \textbf{Residual}
        & \multicolumn{2}{c}{\textbf{ScanQA}}
        & \textbf{SQA3D}
        & \multirow{2}{*}{\textbf{Avg.}} \\
        \textbf{Saliency}  & \textbf{Coverage} & \textbf{Merge} & \textbf{CIDEr} & \textbf{EM} & \textbf{EM} &  \\
        \midrule
        & \checkmark &            & 85.97 & 26.16 & 55.24 & 93.53\% \\
        \checkmark &     &          & 92.04 & 26.50 & 56.81 & 96.99\% \\
        \checkmark & \checkmark &   & 92.67 & 26.95 & 57.57 & 98.19\% \\
        \rowcolor{oursblue}
        \checkmark & \checkmark & \checkmark & 93.95 & 27.51 & 57.63 & \best{99.33\%} \\
        \bottomrule
    \end{tabular}
    }
    \vspace{-10pt}
\end{wraptable}

\vspace{-5pt}
\paragraph{GeoSemZip Component Analysis.}
Table~\ref{tab:projector_component_ablation} evaluates GeoSemZip at a fixed 30\% token budget. Spatial coverage alone retains 93.53\% of performance, while semantic saliency alone reaches 96.99\%. Combining both branches with voxel consolidation raises retention to 98.19\%, supporting the joint use of salient content and spatial coverage. Residual merging adds 1.14 points, reaching 99.33\% without increasing the token budget. This gain highlights the value of aggregating unselected voxel features into contextual anchors rather than discarding them.

\begin{wraptable}{r}{0.47\textwidth}
    \centering
    \caption{ \small
    \textbf{Grouping signal} with anchor layers 9 and 13 and $\kappa=0.5$. Both variants share the same checkpoint, schedule, and decoding protocol and differ only in the ranking used to form the groups.
    }
    \label{tab:grouping_signal}
    \scriptsize
    \setlength{\tabcolsep}{4pt}
    \resizebox{\linewidth}{!}{%
    \begin{tabular}{c|ccc|c}
        \toprule
        \textbf{Grouping} & \multicolumn{2}{c}{\textbf{ScanQA}} & \textbf{SQA3D} 
        & \multirow{2}{*}{\textbf{Avg.}} \\
        \textbf{Signal}   & \textbf{CIDEr}  & \textbf{EM}   & \textbf{EM}   & \\
        \midrule
        Projector-derived   & 85.34 & 25.30 & 55.98 & 92.73\% \\
        \rowcolor{oursblue}
        Query-conditioned   & 88.25 & 26.20 & 56.61 & \best{95.18\%} \\
        \bottomrule
    \end{tabular}%
    }
\end{wraptable}

\paragraph{Grouping Signal.}
Table~\ref{tab:grouping_signal} compares projector-derived and query-conditioned grouping under the same execution schedule. The former reuses the static ordering from GeoSemZip, while the latter ranks visual tokens at each anchor using attention from the final prompt token. Query-conditioned grouping improves all three metrics and raises average performance retention from 92.73\% to 95.18\%. This gain supports updating token priorities according to the question and current LLM layer rather than reusing the projector-derived ordering.

\begin{wraptable}{r}{0.47\textwidth}
    \vspace{-15pt}
  \centering
  \caption{\small
  \textbf{Skipping vs. pruning within the window $[8,23]$.} All variants use the same window-adapted checkpoint without further training and average 638 active visual tokens per LLM layer.
  }
  \label{tab:skip_vs_prune}
  \scriptsize
  \setlength{\tabcolsep}{3pt}
  \resizebox{\linewidth}{!}{%
  \begin{tabular}{lc|ccc|c}
    \toprule
    \multirow{2}{*}{\textbf{Method}}
      & \multirow{2}{*}{\textbf{Anchors}}
      & \multicolumn{2}{c}{\textbf{ScanQA}}
      & \textbf{SQA3D}
      & \multirow{2}{*}{\textbf{Avg.}} \\
      & & \textbf{CIDEr}
      & \textbf{EM}
      & \textbf{EM}
      &  \\
    \midrule
    \multirow{3}{*}{Pruning}
      & 9 & 87.66 & 25.97 & 56.52 & 94.65\% \\
      & 13 & 87.20 & 26.07 & 57.23 & 95.03\% \\
      & 9, 13 & 87.82 & 25.88 & 56.75 & 94.73\% \\
    \rowcolor{oursblue}
    Skipping       & 9, 13 & 88.25 & 26.20 & 56.61 & \best{95.18\%} \\

      \bottomrule
  \end{tabular}%
  }
  \vspace{-5pt}
\end{wraptable}

\paragraph{Skipping vs. Pruning.}
Table~\ref{tab:skip_vs_prune} compares recoverable group-wise skipping with permanent pruning at a matched average token budget. We evaluate pruning at layer 9, layer 13, or both layers. With two anchors, tokens removed at layer 9 remain unavailable at layer 13. At each anchor, skipping selects half of the visual tokens for subsequent computation while preserving the hidden states of the others. At layer 13, all visual tokens are rescored, allowing previously skipped tokens to rejoin the active group. Pruning uses higher keep ratios to match the average number of active visual tokens per LLM layer. Single-anchor pruning retains 94.65\% of performance at layer 9 and 95.03\% at layer 13, while pruning at both anchors retains 94.73\%. Skipping at layers 9 and 13 achieves the highest retention of 95.18\%, outperforming all pruning configurations. These results support the observation in Section~\ref{sec:layer_dynamics} that low token relevance can be temporary and that preserving tokens for later reactivation is beneficial.

\section{Conclusion}
\label{sec:conclusion}

We presented GeoGR, a two-stage compression framework for multi-view 3D VLMs that addresses redundancy in scene representations and visual computation. GeoSemZip consolidates cross-view observations while preserving salient content, spatial coverage, and information from unselected regions. Guided by effective visual computation windows and locally stable but evolving token relevance, GroupRoute combines Late Entry and Early Exit with Recoverable Group-wise Skipping, allowing temporarily inactive tokens to resume computation. A compact, selected dataset supports compression-aware post-training. Experiments demonstrate that GeoGR achieves a favorable performance--efficiency trade-off across 3D VLMs with and without explicit 3D spatial encoding.


\bibliographystyle{plainnat}
\bibliography{EITNLP/references}

\newpage
\appendix
\begingroup
\newsavebox{\appendixtablebox}
\setlength{\textfloatsep}{10pt plus 2pt minus 2pt}
\setlength{\floatsep}{8pt plus 2pt minus 2pt}
\setlength{\intextsep}{6pt plus 2pt minus 1pt}
\setlength{\abovecaptionskip}{4pt}
\setlength{\belowcaptionskip}{5pt}
\renewcommand{\topfraction}{0.90}
\renewcommand{\bottomfraction}{0.80}
\renewcommand{\textfraction}{0.10}
\renewcommand{\floatpagefraction}{0.75}
\brokenpenalty=10000
\makeatletter
\setlength{\@fptop}{0pt}
\setlength{\@fpsep}{12pt}
\setlength{\@fpbot}{0pt plus 1fil}
\setlength{\@dblfptop}{0pt}
\setlength{\@dblfpsep}{12pt}
\setlength{\@dblfpbot}{0pt plus 1fil}
\newcommand{\appendixwrapfinish}{%
  \par
  \ifnum\c@WF@wrappedlines>1
    \begingroup
    \dimen0=\baselineskip
    \multiply\dimen0 by \numexpr\c@WF@wrappedlines-1\relax
    \endgroup
  \fi
  \WFclear
}
\makeatother
\setlength{\columnsep}{12pt}


This appendix supplements the main text with implementation and experimental
details, additional results, and analysis:
\begin{itemize}[leftmargin=1.35em, labelsep=0.45em]
    \item Section~\ref{sec:appendix}: additional experimental details, including
    the training and evaluation protocol and the exact GroupRoute execution
    schedule.
    \item Section~\ref{app:impl}: implementation details, including BARS anchor
    selection.
    \item Section~\ref{app:setup}: experimental setup, including the two 3D VLM
    backbones and the benchmark datasets.
    \item Section~\ref{app:exp}: additional experimental results, including
    post-training data selection, projector-level token-budget comparisons,
    GeoSemZip parameter sensitivity, visual-computation window selection, and
    anchor keep-ratio allocation.
    \item Section~\ref{app:analysis}: additional analysis, including layer-pair
    visual-token ranking and top-200 overlap heatmaps.
\end{itemize}


\section{Additional Experimental Details}
\label{sec:appendix}
\subsection{Training and Evaluation Protocol}
\label{sec:training_protocol}

\paragraph{Input construction.}
For each of the 32 sampled RGB views, we bilinearly pool projector features and vision-encoder attention scores to a $14\times14$ grid. GeoSemZip reconstructs world coordinates from aligned depth maps, camera intrinsics, and camera-to-world poses.
\paragraph{Evaluation.}
Evaluation metrics are summarized in Table~\ref{tab:app_benchmarks}. For ScanQA, predictions and references are lowercased, processed with the benchmark caption-text cleaner, whitespace-normalized, and stripped of trailing punctuation. SQA3D EM follows benchmark-specific normalization: case and punctuation normalization, digit-to-word conversion, article removal, and spelling correction. A prediction is correct if its normalized form matches any normalized reference answer. Evaluation uses batch size 1. 

\paragraph{Post-training.}
For both 3D VLMs, compression-aware post-training consists of projector adaptation followed by late-entry/early-exit adaptation. Each stage runs for one epoch with a global batch size of 16 and learning rate $10^{-5}$. The second stage starts from the projector-adapted checkpoint. The vision encoder remains frozen in both stages; the projector is updated only in the first stage, while the LLM backbone is updated in both. We use zero weight decay, cosine learning rate decay, a warm-up ratio of 0.03, BF16 precision, and DeepSpeed ZeRO-3 in both stages.

\subsection{Exact GroupRoute Execution Schedule in the LLM Decoder}
\label{sec:schedule_cost}

\appendixwrapfinish
\begin{wraptable}{r}{0.52\textwidth}
  \centering
  \caption{\small\textbf{GroupRoute execution schedule on LLaVA-OneVision.}}
  \label{tab:exact_schedule}
  \scriptsize
  \setlength{\tabcolsep}{2.5pt}
  \renewcommand{\arraystretch}{1.05}
  \resizebox{\linewidth}{!}{%
\begin{tabular}{lll}
    \toprule
    \textbf{Blocks} & \textbf{Active sequence} & \textbf{Event} \\
    \midrule
    1--7   & Text only & Visual tokens not yet injected \\
    8      & All & Visual tokens enter \\
    9      & All & Score and group at anchor 9 \\
    10--12 & Text, $\mathcal{H}_{9}$ & $\mathcal{L}_{9}$ bypasses \\
    13     & All & Anchor 13; score and regroup \\
    14--23 & Text, $\mathcal{H}_{13}$ & $\mathcal{L}_{13}$ bypasses \\
    24--28 & Text only & All visual tokens have exited \\
    \bottomrule
  \end{tabular}%
  }
\end{wraptable}

Table~\ref{tab:exact_schedule} summarizes the GroupRoute{} execution schedule
under the adopted LLaVA-OneVision configuration.Visual tokens participate in computation from blocks 8 through 23.The schedule uses grouping anchors at blocks 9 and 13, both with keep ratio
$\kappa=0.5$. An anchor block processes the complete visual token sequence and
recomputes the partition based on its block-input representations, which then governs subsequent blocks until the next anchor.

"All" denotes the full token sequence, including text, visual tokens; $\mathcal{H}_a$ denotes the visual content tokens selected at anchor $a$.

Blocks 8 and 9 both process all retained visual tokens. The anchor-9 partition
governs blocks 10--12, where only $\mathcal{L}_{9}$ bypasses computation; block
13 then processes all visual tokens and forms a new partition. The anchor-13
partition governs blocks 14--23, where only $\mathcal{L}_{13}$ bypasses
computation. Blocks 1--7 and 24--28 contain no visual tokens because of late entry and early
exit.

\appendixwrapfinish
\FloatBarrier
\section{More Implementation Details}
\label{app:impl}

\begingroup
\setlength{\parskip}{3pt}
\setlength{\abovedisplayskip}{5pt plus 1pt minus 1pt}
\setlength{\belowdisplayskip}{5pt plus 1pt minus 1pt}
\setlength{\abovedisplayshortskip}{3pt plus 1pt minus 1pt}
\setlength{\belowdisplayshortskip}{3pt plus 1pt minus 1pt}
\subsection{Budget-Aware Anchor Selection (BARS)}
\label{app:bars}

BARS determines the number and positions of regrouping anchors under fixed GeoSemZip configurations, visual computation windows, and token keep ratios. The algorithm first builds a risk and cost model based on attention scores. It then generates candidate schedules for all feasible anchor numbers and selects the optimal schedule from these candidates following a screening rule.

\paragraph{Query-Guided Visual Token Grouping.}
We sample some of examples from each of the $D=5$ datasets; $\mathcal S_d$ denotes
the examples used from dataset $d$. Inputs follow inference-time frame sampling,
feature ordering, prompt construction, and positional encoding.
The query is the last prompt token for question answering and captioning, or
the grounding token for localization. Let $\mathcal V_x$ contain the positions
of the $N_x$ visual content tokens in the complete prompt. At layer $\ell$, we
compute and normalize their head-averaged attention scores as
\begin{equation*}
\begin{aligned}
s_{x,\ell,i}
&=\frac{1}{H_q}\sum_{h=1}^{H_q}
\operatorname{softmax}\!\left(
\mathbf q_{x,\ell,h}\mathbf K_{x,\ell,h}^{\top}/\sqrt{d_h}
\right)_{i},\\
p_{x,\ell,i}
&=s_{x,\ell,i}\Big/\sum\nolimits_{j\in\mathcal V_x}s_{x,\ell,j},
\qquad i\in\mathcal V_x.
\end{aligned}
\end{equation*}
Here $H_q$ and $d_h$ are the query-head count and head dimension, and
$\mathbf K_{x,\ell,h}$ contains all causally visible prompt keys.Stable descending sorting of $p_{x,\ell,i}$ defines$\operatorname{rank}_{x,\ell}(i)$. At anchor $a$, the top$\lceil\kappa N_x\rceil$ tokens form $\mathcal H_{x,a}$; the rest form$\mathcal L_{x,a}$. Independently ranking layer $\ell$ gives$\mathcal H_{x,\ell}$, used to assess whether the anchor partition has become
stale.

\paragraph{Segment risk.}
Reusing the partition from anchor $a$ at layer $\ell$ incurs three proxy risks:
\begin{equation*}
\begin{aligned}
R_{\mathrm{mass},x}(a,\ell)
  &=\sum\nolimits_{i\in\mathcal L_{x,a}}p_{x,\ell,i},\\
R_{\mathrm{mig},x}(a,\ell)
  &=\frac{|\mathcal L_{x,a}\cap\mathcal H_{x,\ell}|}{|\mathcal H_{x,a}|},\\
R_{\mathrm{rank},x}(a,\ell)
  &=\frac{1}{N_x^2}\sum\nolimits_{i\in\mathcal V_x}
    |\operatorname{rank}_{x,a}(i)-\operatorname{rank}_{x,\ell}(i)|.
\end{aligned}
\end{equation*}
These terms measure the bypassed attention mass, high-priority tokens missed by
the anchor partition, and normalized rank displacement, respectively.

In segment $[a,b]$, block $a$ processes all visual tokens, while blocks
$a+1,\ldots,b$ reuse its partition; $b$ is the next anchor or
$\ell_{\mathrm{out}}+1$. The segment risk is
\begin{equation*}
R_x(a,b)=\sum\nolimits_{\ell=a+1}^{b-1}w(\ell)
\bigl[\alpha R_{\mathrm{mass},x}(a,\ell)
+\beta R_{\mathrm{mig},x}(a,\ell)
+\gamma R_{\mathrm{rank},x}(a,\ell)\bigr],
\end{equation*}
where $w(\ell)=(\ell_{\mathrm{out}}-\ell+1)^p$ assigns more weight to earlier
skipped blocks. We use $\alpha=\beta=0.35$, $\gamma=0.2$, and $p=1$.
For each dataset, we average over its sampled examples,
$R_d(a,b)=|\mathcal S_d|^{-1}\sum_{x\in\mathcal S_d}R_x(a,b)$, and then average
equally across datasets, $R(a,b)=D^{-1}\sum_{d=1}^{D}R_d(a,b)$.

\paragraph{Visual token--layer budget.}
Let $C_x^{\mathrm{full}}$ and $C_x^{\mathrm{group}}$ denote the actual numbers of
active visual-sequence positions in a full and grouped block for example $x$.
For schedule
$\mathcal A=\{a_1<\cdots<a_m\}$ over the window
$[\ell_{\mathrm{in}},\ell_{\mathrm{out}}]$, the prefix before $a_1$ and the
$m$ anchors execute all visual tokens; the remaining blocks use the groups.
Thus,
\begin{equation*}
C_x(\mathcal A)
=(a_1-\ell_{\mathrm{in}}+m)C_x^{\mathrm{full}}
+(\ell_{\mathrm{out}}-a_1+1-m)C_x^{\mathrm{group}}.
\end{equation*}
We average costs within and across datasets, then normalize by the corresponding
full-window cost $(\ell_{\mathrm{out}}-\ell_{\mathrm{in}}+1)C_x^{\mathrm{full}}$
to obtain $\Gamma(\mathcal A)$.

\begingroup
\floatstyle{ruled}
\restylefloat{algorithm}
\begin{algorithm}[!t]
\small
\caption{Budget-constrained Anchor and Regrouping Search (BARS)}
\label{alg:bars}
\begin{algorithmic}[1]
\Require Sampled examples $\{\mathcal S_d\}_{d=1}^{D}$; fixed checkpoint and GeoSemZip configuration
\Require Visual window $\mathcal W$, keep ratio $\kappa$, budget $B$, anchor limit $M_{\max}$, gap $g$, threshold $\tau$
\Ensure Selected anchor schedule $\mathcal A^*$
\ForAll{sampled examples and layers in $\mathcal W$}
\State \hspace*{1em}Collect query-to-visual attention; normalize over visual content tokens
\State \hspace*{1em}Store stable token ranks and the high- and low-priority groups
\EndFor
\ForAll{legal segments $(a,b)$}
\State \hspace*{1em}Evaluate attention mass, group migration, and rank displacement on intervening blocks
\State \hspace*{1em}Apply depth weights; average within datasets to obtain $\{R_d(a,b)\}$
\State \hspace*{1em}Average across datasets to obtain $R(a,b)$; compute the segment cost
\EndFor
\State Initialize DP states for possible first anchors with full-prefix costs and zero risk
\ForAll{reachable DP states in increasing boundary order}
\ForAll{legal next segments}
\State \hspace*{2em}Extend the state with the segment cost, risk, and one additional anchor
\If{the extended state satisfies the budget and anchor limit}
\State \hspace*{3em}Keep the lowest-risk predecessor for its boundary, cost, and anchor count
\EndIf
\EndFor
\EndFor
\For{each feasible anchor count $m=1,\ldots,M_{\max}$}
\State \hspace*{1em}Backtrack the lowest-risk terminal state to obtain $\mathcal A_m$
\State \hspace*{1em}Sum its segment risks for each dataset to obtain $\{R_d(\mathcal A_m)\}$
\EndFor
\State $\mathcal A^*\gets\mathcal A_1$
\For{each larger feasible anchor count $m$, in increasing order}
\If{$R(\mathcal A_m)<R(\mathcal A^*)$ and every dataset's relative risk reduction is at least $\tau$}
\State \hspace*{2em}$\mathcal A^*\gets\mathcal A_m$
\EndIf
\EndFor
\State \textbf{return} $\mathcal A^*$
\end{algorithmic}
\end{algorithm}
\endgroup

\paragraph{Candidate schedules by dynamic programming.}
A schedule $\mathcal A=(a_1<\cdots<a_m)$ is scored by its total segment risk
$\sum_{j=1}^{m}R(a_j,a_{j+1})$, where $a_{m+1}=\ell_{\mathrm{out}}+1$ closes the
last segment and the blocks before $a_1$ carry cost but no skipping risk. The
cost is simpler than the risk: the full-visual blocks are the
$a_1-\ell_{\mathrm{in}}$ prefix blocks and the $m$ anchors, so $\Gamma$ is
determined by $a_1$ and $m$ alone. The budget is therefore a closed-form
condition on $(a_1,m)$ instead of a quantity accumulated along the schedule, and
for each anchor count $m\leq M_{\max}$ the search minimizes risk over a feasible
set $\mathcal F_m$:
\begin{equation*}
\mathcal A_m=\operatorname*{arg\,min}_{\mathcal A\in\mathcal F_m}
\sum_{j=1}^{m}R(a_j,a_{j+1}),
\qquad
\mathcal F_m=\bigl\{\mathcal A:\ |\mathcal A|=m,\
a_{j+1}-a_j\geq g,\ \Gamma(a_1,m)\leq B\bigr\}.
\end{equation*}
Let $V(\ell,j)$ denote the smallest accumulated risk of placing $j$ anchors with
the last one at block $\ell$. With the segment risks precomputed, $V(a_1,1)=0$
for every budget-feasible $a_1$, and
\begin{equation*}
V(\ell',j+1)=\min_{\ell\leq\ell'-g}\bigl[V(\ell,j)+R(\ell,\ell')\bigr].
\end{equation*}
Minimizing $V(\ell,m)+R(\ell,\ell_{\mathrm{out}}+1)$ over the last anchor and
backtracking gives $\mathcal A_m$ and its per-dataset risk
$R_d(\mathcal A_m)=\sum_jR_d(a_j,a_{j+1})$. The window spans at most 17 blocks
and $M_{\max}=4$.

\paragraph{Selecting the anchor count.}
Starting from the feasible one-anchor candidate $\mathcal A_1$, BARS examines
larger feasible counts in increasing order. A candidate $\mathcal A_m$ replaces
the current selection $\mathcal A^*$ only when
\begin{equation*}
R(\mathcal A_m)<R(\mathcal A^*),\qquad
R_d(\mathcal A_m)\leq(1-\tau)R_d(\mathcal A^*)\quad\text{for every }d,
\end{equation*}
where $R(\mathcal A)=D^{-1}\sum_dR_d(\mathcal A)$ and $\tau=5\%$.

Both 3D VLMs share the search parameters: keep ratio $\kappa=0.5$, minimum gap
$g=2$, anchor limit $M_{\max}=4$, and margin $\tau=5\%$, with no maximum-span
constraint. They differ only in the visual computation window, $[8,23]$ for 
LLaVA-OneVision and $[8,24]$ for Video-3D LLM. The optimal anchor schedule for 
both 3D VLMs is $[9,13]$. The schedule is fixed at inference, while the token 
groups are recomputed for each query. Algorithm~\ref{alg:bars} summarizes the 
overall procedure.

\endgroup


\appendixwrapfinish
\FloatBarrier
\section{More Experimental Setup Details}
\label{app:setup}

\subsection{3D VLMs}
\label{app:3dvlm}
\paragraph{LLaVA-OneVision-7B.} LLaVA-OneVision~\citep{li2024llavaov} provides a unified framework for single-image, multi-image, and video understanding. Its SigLIP visual encoder extracts patch-level visual features, which are further projected into the language embedding space via a two-layer MLP. Each sampled scene view is independently encoded, and the output visual tokens are concatenated with textual instructions to form the multimodal input sequence. Under our 32-view setting, each view yields a $14\times14$ pooled visual grid, producing $32\times14\times14=6{,}272$ visual content tokens before compression.
This 3D VLM takes only RGB-based visual features as input, while depth maps and camera calibration parameters are exclusively fed into our GeoSemZip{} module.
\paragraph{Video-3D LLM-7B.} Built upon LLaVA-Video-7B, Video-3D LLM~\citep{zheng2024video3dllm} explicitly integrates 3D positional information into multi-view visual representations. Specifically, depth maps, camera intrinsics, and camera poses project observations from different views into a unified world coordinate system. The obtained patch-level spatial coordinates are encoded and fused with visual features, endowing the model with robust spatial awareness.
In addition to standard visual question answering, Video-3D LLM supports spatial reasoning tasks, including object-conditioned captioning and visual grounding. For object-conditioned captioning, the center coordinates of target objects are adopted to specify query instances. For visual grounding, the hidden representations of a dedicated grounding token are matched with object-level features to locate target objects. 

\begin{table}[!htbp]
  \caption{\small\textbf{Architectures and scene inputs of the evaluated 3D VLMs.}
  Token counts use the 32-view setting and exclude formatting tokens. Q/KV
  denotes query and key/value attention heads.}
  \label{tab:app_backbones}
  \centering
  \small
  \setlength{\tabcolsep}{2pt}
  \renewcommand{\arraystretch}{1.08}
  \begin{tabular}{@{}>{\raggedright\arraybackslash}p{0.27\linewidth}>{\raggedright\arraybackslash}p{0.26\linewidth}>{\raggedright\arraybackslash}p{\dimexpr0.47\linewidth-6\tabcolsep\relax}@{}}
    \toprule
    \textbf{Component} & \textbf{LLaVA-OneVision-7B} & \textbf{Video-3D LLM-7B} \\
    \midrule
    Pretrained VLM family & LLaVA-OneVision & LLaVA-Video \\
    Vision encoder & SigLIP SO400M/14 & SigLIP SO400M/14 \\
    Encoder input resolution & $384\times384$ & $384\times384$ \\
    Visual feature dimension & 1,152 & 1,152 \\
    Multimodal projector & Two-layer MLP, GELU & Two-layer MLP, GELU \\
    LLM & Qwen2-7B & Qwen2-7B \\
    Decoder blocks & 28 & 28 \\
    Attention heads (Q/KV) & 28 / 4 & 28 / 4 \\
    Hidden / FFN dimension & 3,584 / 18,944 & 3,584 / 18,944 \\
    \midrule
    Scene view sampling & 32 uniform RGB views & 32 uniform RGB views \\
    Pooled grid per view & $14\times14$ & $14\times14$ \\
    Uncompressed content tokens & 6,272 & 6,272 \\
    Role of depth and poses & Geometric compression & 3D encoding and compression \\
    Evaluated task interfaces & Question answering & Question answering, captioning, grounding \\
    \bottomrule
  \end{tabular}
\end{table}

\subsection{Benchmarks and Datasets}
\label{app:bench}
The five ScanNet-based datasets serve both as sources of post-training examples
and as benchmarks for 3D question answering, object captioning, and
language grounding. Table~\ref{tab:app_benchmarks} reports, for each dataset,
the number of examples in our full post-training pool before PRISM selection
and the separate split used for evaluation. The full pool contains 223,128
examples across the five sources; its selected subset is reported in
Table~\ref{tab:posttrain_data}. LLaVA-OneVision is evaluated on the two
question-answering benchmarks, whereas Video-3D LLM is evaluated on all five.

\begin{table}[!htbp]
  \caption{\small\textbf{Training sources and evaluation splits for five ScanNet-based benchmarks.}
  Train counts examples before PRISM selection; Eval counts held-out examples.
  Evaluation and scene counts follow the Video-3D LLM annotations~\citep{zheng2024video3dllm}.}
  \label{tab:app_benchmarks}
  \centering
  \footnotesize
  \setlength{\tabcolsep}{4pt}
  \renewcommand{\arraystretch}{1.08}
  \resizebox{\linewidth}{!}{%
  \begin{tabular}{@{}l|rlrr|l@{}}
    \toprule
    \textbf{Benchmark} & \textbf{Train} & \textbf{Eval split}
      & \textbf{Eval} & \textbf{Scenes} & \textbf{Reported metrics} \\
    \midrule
    ScanQA~\citep{azuma2022scanqa} & 26,515 & Validation & 4,675 & 71 & CIDEr, normalized EM \\
    SQA3D~\citep{ma2023sqa3d} & 79,445 & Test & 3,519 & 67 & Normalized EM \\
    Scan2Cap~\citep{chen2021scan2cap} & 36,665 & Validation & 2,068 & 141 & BLEU-4, CIDEr (IoU 0.5) \\
    ScanRefer~\citep{chen2020scanrefer} & 36,665 & Validation & 9,508 & 141 & Acc@0.25, Acc@0.5 \\
    Multi3DRefer~\citep{zhang2023multi3drefer} & 43,838 & Validation & 11,120 & 141 & F1@0.25, F1@0.5 \\
    \bottomrule
  \end{tabular}}
\end{table}

\paragraph{ScanQA.}
ScanQA~\citep{azuma2022scanqa} contains human-written questions and answers
grounded in indoor 3D scenes. Questions require identifying objects, recognizing
their attributes, and reasoning about their spatial relationships.We report CIDEr for generated-answer agreement and normalized exact match for answer correctness on the validation split.

\paragraph{SQA3D.}
SQA3D~\citep{ma2023sqa3d} adds a textual description of an agent's situation,
including its position and orientation, to a question about the surrounding
scene. Answering requires interpreting the question from that situated
perspective. The benchmark therefore tests whether the compressed representation
preserves scene layout and spatial relations as well as object appearance.
We evaluate the test split using normalized EM, with the situation description
and question supplied together in the task prompt. 

\paragraph{Scan2Cap.}
Scan2Cap~\citep{chen2021scan2cap} evaluates dense captioning in 3D scenes,
using object descriptions from ScanRefer. A caption must describe the localized
object and may refer to its attributes and surrounding objects. For Video-3D, the target location is supplied through its coordinate-conditioned captioning prompt. We report BLEU-4 and CIDEr at a 3D IoU threshold of 0.5,which couples language quality with adequate object localization. 

\paragraph{ScanRefer.}
ScanRefer~\citep{chen2020scanrefer} asks a model to localize one object in a
3D scene from a natural-language referring expression. Descriptions use
appearance and relational cues to distinguish the target from nearby objects.
We report the fraction of predictions whose 3D bounding-box intersection over
union with the target reaches 0.25 or 0.5, denoted Acc@0.25 and Acc@0.5.

\paragraph{Multi3DRefer.}
Multi3DRefer~\citep{zhang2023multi3drefer} extends language grounding to
expressions that may refer to zero, one, or multiple objects. A model must
identify the appropriate set of objects and handle descriptions with distractors
or no valid target. We use the benchmark's set-level F1 metrics at 3D IoU
thresholds of 0.25 and 0.5, including its zero-target evaluation convention.

\FloatBarrier
\section{Additional Experimental Results}
\label{app:exp}

\subsection{Effect of Data Selection on Post-Training Performance}
\label{app:posttrain_results}

Table~\ref{tab:app_posttrain_results} provides per-metric evaluation results for our data selection ablation study. For each 3D VLM, we compare two sets of dense checkpoints: one post-trained on the complete five-source dataset, and the other post-trained on the PRISM-selected subset (approximately 30\% from each source). Retention is defined as the unweighted mean of the ratios between the results obtained on the selected subset and those on the full dataset across evaluation metrics of the corresponding model.
Our results show that using this 30\% subset maintains strong overall performance: Video-3D LLM achieves 98.72\% performance retention, while LLaVA-OneVision retains 95.69\% of the full-data post-trained performance.

\begin{table}[!htb]
  \caption{\small\textbf{Dense post-training on full data and PRISM-selected data.}
  PRISM retains approximately 30\% from each source; retention averages the
  metric-wise ratios to the full-data post-trained checkpoint.}
  \label{tab:app_posttrain_results}
  \centering
  \small
  \setlength{\tabcolsep}{4pt}
  \renewcommand{\arraystretch}{1.08}
  \textbf{(a) LLaVA-OneVision-7B}\par\smallskip
  \begin{tabular}{l|rrr|r}
    \toprule
    \textbf{Training data} & \textbf{ScanQA CIDEr} & \textbf{ScanQA EM}
      & \textbf{SQA3D EM} & \textbf{Retention (\%)} \\
    \midrule
    \rowcolor{gray!20}
    Full set (223,128) & 100.27 & 29.07 & 59.53 & 100.00 \\
    PRISM 30\% (66,940) & 94.84 & 28.09 & 57.06 & 95.69 \\
    \bottomrule
  \end{tabular}

\medskip
  \textbf{(b) Video-3D LLM-7B}\par\smallskip
  \scriptsize
  \setlength{\tabcolsep}{2pt}
  \resizebox{\linewidth}{!}{%
  \begin{tabular}{l|rrrrrrrrr|r}
    \toprule
    \multirow{2}{*}{\textbf{Training data}}
      & \multicolumn{2}{c}{\textbf{ScanQA}}
      & \textbf{SQA3D}
      & \multicolumn{2}{c}{\textbf{Scan2Cap}}
      & \multicolumn{2}{c}{\textbf{ScanRefer}}
      & \multicolumn{2}{c}{\textbf{Multi3DRefer}}
      & \multirow{2}{*}{\textbf{Retention (\%)}} \\
    \cmidrule(lr){2-3}\cmidrule(lr){5-6}\cmidrule(lr){7-8}\cmidrule(lr){9-10}
      & \textbf{CIDEr} & \textbf{EM} & \textbf{EM}
      & \textbf{B-4@.5} & \textbf{C@.5}
      & \textbf{Acc@.25} & \textbf{Acc@.5}
      & \textbf{F1@.25} & \textbf{F1@.5} & \\
    \midrule
    \rowcolor{gray!20}
    Full set (223,128)
      & 99.51 & 28.72 & 57.71 & 39.12 & 80.66
      & 59.88 & 53.39 & 57.59 & 52.50 & 100.00 \\
    PRISM 30\% (66,940)
      & 98.19 & 27.74 & 57.43 & 40.71 & 80.02
      & 57.79 & 51.48 & 56.95 & 51.75 & 98.72 \\
    \bottomrule
  \end{tabular}}
\end{table}

\subsection{GeoSemZip Budget Selection and Baseline Comparison}
\label{sec:projector_results}
\label{sec:projector_full_sweeps}

\appendixwrapfinish
\begin{wraptable}{r}{0.48\textwidth}
  \centering
  \vspace{-13pt}
  \caption{\small\textbf{GeoSemZip budget selection on LLaVA-OneVision.}
  Each budget uses a separately post-trained checkpoint. Vis. tokens is the
  ScanQA mean entering the LLM; Avg. is retention relative to Dense.}
  \label{tab:geosemzip_budget_choice}
  \small
  \setlength{\tabcolsep}{5pt}
  \renewcommand{\arraystretch}{1.06}
  \begin{tabular}{@{}lrr@{}}
    \toprule
    \textbf{Budget} & \textbf{Vis. tokens} & \textbf{Avg. (\%)} \\
    \midrule
    $\approx 20\%$ & 1255.00 & 97.26 \\
    \rowcolor{oursblue}
    $\approx 30\%$ & 1881.73 & \best{99.33} \\
    $\approx 40\%$ & 2475.47 & 98.38 \\
    \bottomrule
  \end{tabular}
  \vspace{-15pt}
\end{wraptable}

We examine GeoSemZip at 20\%, 30\%, and 40\% to select the Stage-I budget
(Table~\ref{tab:geosemzip_budget_choice}). We compare post-trained
projector-level compression methods on LLaVA-OneVision at target budgets of
approximately 10\%, 20\%, and 30\% of the original 6,272 visual tokens
(Table~\ref{tab:projector_budget}).

\paragraph{Choice of the 30\% budget.}
Table~\ref{tab:geosemzip_budget_choice} isolates GeoSemZip's Stage-I
accuracy--token trade-off. The 30\% setting retains 99.33\% of Dense performance
with a mean of 1,881.73 visual tokens, compared with 97.26\% retention at 20\%.
Increasing the budget to 40\% requires 593.74 more tokens and yields a lower
retention of 98.38\%. We therefore use the post-trained 30\% checkpoint as the
input to GroupRoute.
\appendixwrapfinish

\begin{table}[!t]
  \centering
  \small
  \setlength{\tabcolsep}{7pt}
  \renewcommand{\arraystretch}{1.08}
  \sbox{\appendixtablebox}{%
  \begin{tabular}{l|rrrr|r}
    \toprule
    \multirow{2}{*}{\textbf{Method}}
      & \multirow{2}{*}{\textbf{Vis. tokens}}
      & \multicolumn{2}{c}{\textbf{ScanQA}}
      & \textbf{SQA3D}
      & \multirow{2}{*}{\textbf{Avg. (\%)}} \\
    \cmidrule(lr){3-4}\cmidrule(lr){5-5}
      & & \textbf{CIDEr} & \textbf{EM} & \textbf{EM} & \\
    \midrule
    \rowcolor{gray!20}
    Dense & 6272.00 & 94.84 & 28.09 & 57.06 & 100.00 \\
    \midrule
    \rowcolor{lightgray}
    \multicolumn{6}{c}{\emph{Token budget $\approx 10\%$}} \\
    VisionZip & 640.00 & 81.36 & 23.59 & 54.33 & 88.34 \\
    VisPruner & 640.00 & 84.67 & 25.09 & 56.01 & 92.26 \\
    SeGPruner & 640.00 & 84.52 & 24.86 & 55.61 & 91.69 \\
    DTC & 702.00 & 86.06 & 25.50 & 54.96 & 92.61 \\
    \rowcolor{oursblue}
    GeoSemZip{} (ours) & 628.00 & 87.44 & 25.41 & 56.69 & \best{94.00} \\
    \midrule
    \rowcolor{lightgray}
    \multicolumn{6}{c}{\emph{Token budget $\approx 20\%$}} \\
    VisionZip & 1248.00 & 88.10 & 25.97 & 56.12 & 94.57 \\
    VisPruner & 1248.00 & 86.29 & 25.24 & 56.78 & 93.45 \\
    SeGPruner & 1248.00 & 87.43 & 25.73 & 57.26 & 94.71 \\
    DTC & 1285.00 & 87.99 & 25.41 & 57.00 & 94.38 \\
    \rowcolor{oursblue}
    GeoSemZip{} (ours) & 1255.00 & 90.01 & 26.78 & 57.94 & \best{97.26} \\
    \midrule
    \rowcolor{lightgray}
    \multicolumn{6}{c}{\emph{Token budget $\approx 30\%$}} \\
    VisionZip & 1856.00 & 91.27 & 27.40 & 56.86 & 97.81 \\
    VisPruner & 2048.00 & 90.03 & 26.52 & 57.09 & 96.46 \\
    SeGPruner & 1888.00 & 90.34 & 26.87 & 57.23 & 97.07 \\
    DTC & 1756.67 & 93.25 & 27.25 & 56.72 & 98.25 \\
    \rowcolor{oursblue}
    GeoSemZip{} (ours) & 1881.73 & 93.95 & 27.51 & 57.63 & \best{99.33} \\
    \bottomrule
  \end{tabular}%
  }
  \begin{minipage}{\wd\appendixtablebox}
  \caption{\small\textbf{GeoSemZip and projector-level baselines across token budgets on LLaVA-OneVision.}
  All reported configurations are post-trained. Vis. tokens is the ScanQA mean
  entering the LLM; Avg. is retention relative to Dense. \best{Bold} marks the
  highest reported Avg. within each budget group.}
  \label{tab:projector_budget}
  \label{tab:projector_budget_main}
  \centering
  \usebox{\appendixtablebox}
  \end{minipage}
\end{table}

\paragraph{Comparison with baselines.}
At the approximately 10\%, 20\%, and 30\% target budgets in
Table~\ref{tab:projector_budget}, GeoSemZip achieves the highest average
retention, exceeding the strongest baseline in each group by 1.39, 2.55, and
1.08 percentage points, respectively. The results show that its combination
of salient-token selection and spatially distributed anchors is effective
across these tested budgets.
\appendixwrapfinish

\appendixwrapfinish
\subsection{GeoSemZip Parameter Sensitivity}
\appendixwrapfinish
\begin{wraptable}{R}{0.50\textwidth}
  \centering
  \caption{\small\textbf{GeoSemZip voxel-size ablation at a 30\% token budget.}
  LLaVA-OneVision variants are post-trained independently; $K$ is the mean
  number of occupied voxels per scene.}
  \label{tab:projector_voxel_size}
  \scriptsize
  \setlength{\tabcolsep}{2.5pt}
  \renewcommand{\arraystretch}{1.05}
  \resizebox{\linewidth}{!}{%
\begin{tabular}{cc|c|rrr|r}
    \toprule
    \multirow{2}{*}{\makecell{\textbf{Voxel}\\\textbf{size $s$}}}
      & \multirow{2}{*}{$K$}
      & \multirow{2}{*}{\makecell{\textbf{Keep}\\\textbf{ratio}}}
      & \multicolumn{2}{c}{\textbf{ScanQA}}
      & \multicolumn{1}{c|}{\textbf{SQA3D}}
      & \multirow{2}{*}{\textbf{Avg.}} \\
    \cmidrule(lr){4-5}\cmidrule(lr){6-6}
      & & & \textbf{CIDEr} & \textbf{EM} & \textbf{EM} & \\
    \midrule
    0.05 & 4778.6 & 30\% & 92.10 & 27.40 & 57.26 & 98.34\% \\
    \rowcolor{oursblue}
    0.10 & 3574.0 & 30\% & 93.95 & 27.51 & 57.63 & \best{99.33\%} \\
    0.15 & 2520.8 & 30\% & 91.31 & 27.42 & 57.63 & 98.30\% \\
    \bottomrule
  \end{tabular}%
  }
\end{wraptable}

\paragraph{Voxel size.}
The voxel width $s$ sets the granularity of geometric consolidation. A small $s$
keeps many distinct voxels and leaves part of the cross-view duplication
unresolved, whereas a large $s$ merges more aggressively and starts to fuse
surfaces that belong to different objects. Table~\ref{tab:projector_voxel_size}
sweeps $s$ at the same 30\% budget.The default $s=0.1$, which leaves 3574 occupied voxels per scene on average,gives the best trade-off between removing duplicates and preserving object
boundaries.

\appendixwrapfinish
\begin{wraptable}{R}{0.50\textwidth}
  \centering
  \vspace{-5pt}
  \caption{\small\textbf{Voxel-score aggregation for GeoSemZip at a 30\% token budget.}
  LLaVA-OneVision variants are post-trained independently and differ only in
  the within-voxel score-pooling rule.}
  \label{tab:projector_aggregation}
  \scriptsize
  \setlength{\tabcolsep}{2.5pt}
  \renewcommand{\arraystretch}{1.05}
  \resizebox{\linewidth}{!}{%
\begin{tabular}{l|rrr|r}
    \toprule
    \multirow{2}{*}{\textbf{Aggregation}}
      & \multicolumn{2}{c}{\textbf{ScanQA}}
      & \multicolumn{1}{c}{\textbf{SQA3D}}
      & \multirow{2}{*}{\textbf{Avg. (\%)}} \\
    \cmidrule(lr){2-3}\cmidrule(lr){4-4}
      & \textbf{CIDEr} & \textbf{EM} & \textbf{EM} & \\
    \midrule
    Mean & 93.15 & 27.76 & 57.12 & 99.06 \\
    \rowcolor{oursblue}
    Max (default) & 93.95 & 27.51 & 57.63 & \best{99.33} \\
    \bottomrule
  \end{tabular}%
  }
\end{wraptable}

\newpage
\paragraph{Score aggregation.}
A voxel inherits a single score from the patches it contains, and the pooling rule
decides whose score that is. Max pooling lets the best-scoring view of a region
determine its rank, whereas mean pooling dilutes that view with the others, which
matters because a region is often salient in only a few of the views that observe
it. Table~\ref{tab:projector_aggregation} confirms the effect: mean pooling costs
0.27 points of Avg.\ at the same token budget, so we keep max pooling.

\appendixwrapfinish
\begin{wraptable}{r}{0.50\textwidth}
  \centering
  \vspace{-10pt}
  \caption{\small\textbf{GeoSemZip budget-split ablation at a 30\% token budget.}
  On LLaVA-OneVision, independently post-trained variants differ in $\rho$,
  the fraction assigned to dominant tokens; residual merging remains enabled.}
  \label{tab:projector_rho}
  \scriptsize
  \setlength{\tabcolsep}{2.5pt}
  \renewcommand{\arraystretch}{1.05}
  \resizebox{\linewidth}{!}{%
\begin{tabular}{l|rrr|r}
    \toprule
    \multirow{2}{*}{\textbf{Dominant ratio $\rho$}}
      & \multicolumn{2}{c}{\textbf{ScanQA}}
      & \multicolumn{1}{c}{\textbf{SQA3D}}
      & \multirow{2}{*}{\textbf{Avg. (\%)}} \\
    \cmidrule(lr){2-3}\cmidrule(lr){4-4}
      & \textbf{CIDEr} & \textbf{EM} & \textbf{EM} & \\
    \midrule
    0.65 & 91.46 & 27.40 & 57.46 & 98.23 \\
    0.75 & 92.89 & 27.66 & 57.37 & 98.99 \\
    \rowcolor{oursblue}
    0.85 (default) & 93.95 & 27.51 & 57.63 & \best{99.33} \\
    0.95 & 92.36 & 27.61 & 57.00 & 98.52 \\
    \bottomrule
  \end{tabular}%
  }
\end{wraptable}

\paragraph{Budget Split.}
The ratio $\rho$ partitions the budget between semantically salient dominant tokens and spatially distributed contextual anchors. Since the total token budget is independent of $\rho$, all configurations in Table~\ref{tab:projector_rho} process the same number of tokens and only differ in token allocation. We observe performance degradation when $\rho$ deviates from the optimal value in either direction. Reducing $\rho$ to 0.75 and 0.65 decreases the average metric by 0.34 and 1.10 points, respectively, while increasing $\rho$ to 0.95 incurs a 0.81-point drop. This indicates that most of the budget should be assigned to salient features, with a small yet non-negligible portion reserved to maintain full scene coverage.


\appendixwrapfinish
\subsection{Visual-Computation Window Selection}

\begin{wraptable}{R}{0.53\textwidth}
  \centering
  \vspace{-10pt}
  \caption{\small\textbf{Visual-computation window selection on LLaVA-OneVision.}
  We first vary late entry with visual tokens retained through block 28, then
  vary early exit with entry fixed at block 8. Avg. is relative to the dense
  reference in Table~\ref{tab:main}.}
  \label{tab:window_ablation}
  \scriptsize
  \setlength{\tabcolsep}{2.5pt}
  \renewcommand{\arraystretch}{1.05}
  \resizebox{\linewidth}{!}{%
  \begin{tabular}{lrrr|r}
    \toprule
    \multirow{2}{*}{\textbf{Window}}
      & \multicolumn{2}{c}{\textbf{ScanQA}}
      & \multicolumn{1}{c|}{\textbf{SQA3D}}
      & \multirow{2}{*}{\textbf{Avg.}} \\
    \cmidrule(lr){2-3}\cmidrule(lr){4-4}
      & \textbf{CIDEr} & \textbf{EM} & \textbf{EM} & \\
    \midrule
    \rowcolor{lightgray}
    \multicolumn{5}{c}{\emph{Step 1: late entry (visual tokens through block 28)}} \\
    $[5,28]$ & 87.60 & 25.56 & 56.27 & 93.99\% \\
    $[6,28]$ & 87.87 & 25.13 & 56.15 & 93.51\% \\
    $[7,28]$ & 85.37 & 25.41 & 55.61 & 92.65\% \\
    \rowcolor{oursblue}
    $[8,28]$ & 86.61 & 25.35 & 56.49 & 93.53\% \\
    $[9,28]$ & 86.88 & 25.33 & 55.98 & 93.30\% \\
    \midrule
    \rowcolor{lightgray}
    \multicolumn{5}{c}{\emph{Step 2: early exit (late entry at block 8)}} \\
    $[8,22]$ & 83.38 & 24.79 & 55.19 & 90.97\% \\
    \rowcolor{oursblue}
    $[8,23]$ & 88.13 & 26.25 & 57.00 & \best{95.43\%} \\
    $[8,24]$ & 84.86 & 24.94 & 55.87 & 92.06\% \\
    $[8,28]$ & 86.61 & 25.35 & 56.49 & 93.53\% \\
    \bottomrule
  \end{tabular}%
  }
\end{wraptable}

The KL interventions in Figure~\ref{fig:visual_window_kl} guide the candidate
boundaries before post-training. Skipping the first $K=5$--$7$ blocks gives the
last consecutive low-divergence region for late entry, corresponding to entry
near blocks 6--8. For early exit, divergence drops between retaining visual
tokens through blocks 21 and 22 and remains low from block 23 onward. We then
compare nearby windows using the same input and post-training protocol.

We determine the two boundaries sequentially. First, we compare candidate late-entry layers without early exit. Entry at block 8 achieves an average performance of 93.53\%, which is only 0.46 percentage points lower than the optimal entry scheme at block 5, while reducing visual computation for three additional blocks. We therefore fix the late-entry position at block 8. With the late-entry position fixed, we then compare candidate early-exit layers. Exiting after block 23 yields the highest average performance (95.43\%), surpassing the schemes that exit after blocks 22 and 24 by 4.46 and 3.37 percentage points, respectively. The resulting computation window is $[8,23]$, and we adopt the post-trained checkpoint of this window for group-wise skipping experiments on LLaVA-OneVision.
\appendixwrapfinish

\appendixwrapfinish
\subsection{Keep-Ratio Allocation Across Anchors}
\label{sec:keep_ratio_allocation}

\appendixwrapfinish
\begin{wraptable}{R}{0.52\textwidth}
  \centering
  \vspace{-10pt}
  \caption{\small\textbf{Keep-ratio allocation at GroupRoute anchors 9 and 13.}
  On LLaVA-OneVision with window $[8,23]$, ``Visual tokens'' is the mean
  active count per LLM layer; Avg. is relative to the dense model.}
  \label{tab:keep_ratio_alloc}
  \scriptsize
  \setlength{\tabcolsep}{2.5pt}
  \renewcommand{\arraystretch}{1.05}
  \resizebox{\linewidth}{!}{%
\begin{tabular}{lr|rrr|r}
    \toprule
    \multirow{2}{*}{\textbf{Keep ratios (9\,/\,13)}}
      & \multirow{2}{*}{\makecell{\textbf{Visual}\\\textbf{tokens}}}
      & \multicolumn{2}{c}{\textbf{ScanQA}}
      & \multicolumn{1}{c|}{\textbf{SQA3D}}
      & \multirow{2}{*}{\textbf{Avg.}} \\
    \cmidrule(lr){3-4}\cmidrule(lr){5-5}
      & & \textbf{CIDEr} & \textbf{EM} & \textbf{EM} & \\
    \midrule
    0.75\,/\,0.25 & 520.79 & 87.28 & 25.99 & 56.38 & 94.46\% \\
    \rowcolor{oursblue}
    0.50\,/\,0.50 (default) & 638.38 & 88.25 & 26.20 & 56.61 & \best{95.18\%} \\
    0.25\,/\,0.75 & 755.98 & 87.77 & 26.07 & 56.07 & 94.55\% \\
    \bottomrule
  \end{tabular}%
  }
\end{wraptable}

At each anchor $a$, the keep ratio $\kappa_a$ specifies the fraction of visual
tokens assigned to the high-priority group. The remaining tokens form the
low-priority group and skip computation in the subsequent non-anchor blocks.
We compare three pairs of ratios $(\kappa_9,\kappa_{13})$. The groups
formed at blocks 9 and 13 govern blocks 10--12 and 14--23, respectively;
therefore, the two ratios affect different numbers of blocks and lead to
different computational costs.
Among the tested allocations, equal keep ratios $(0.50,0.50)$ yield the highest
average retention (95.18\%). The $(0.75,0.25)$ setting uses fewer active visual
tokens but retains 94.46\% of performance; reversing the ratios increases the
active-token count while retaining 94.55\%. We therefore use
$\kappa_9=\kappa_{13}=0.50$.


\appendixwrapfinish
\FloatBarrier
\section{Additional Analysis}
\label{app:analysis}
\subsection{Layer-Pair visual token Relevance Stability}
\label{app:ranking_heatmaps}

\begin{figure}[!htbp]
  \centering
  \begin{minipage}[t]{0.48\textwidth}
    \centering
    \includegraphics[width=\linewidth,trim=0 0 0 6bp,clip]{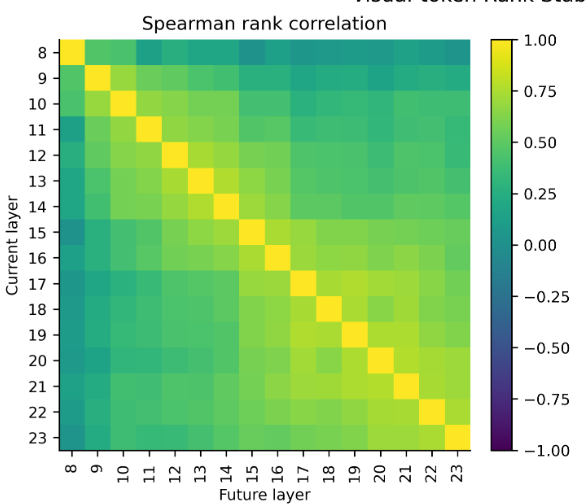}
    \par\smallskip\textbf{(a)} Spearman rank correlation
  \end{minipage}\hfill
  \begin{minipage}[t]{0.48\textwidth}
    \centering
    \includegraphics[width=\linewidth]{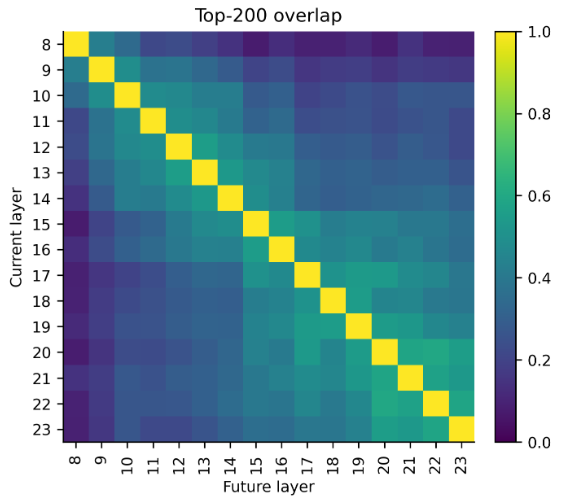}
    \par\smallskip\textbf{(b)} Top-200 overlap
  \end{minipage}
  \caption{\small\textbf{Layer-pair stability of visual-token relevance on ScanQA.}
  For LLaVA-OneVision decoder layers 8--23, (a) gives Spearman correlation
  between complete visual-token rankings, and (b) gives the fraction shared
  by the two layers' top-200 sets.}
  \label{fig:app_ranking_heatmap}
\end{figure}

We examine visual-token relevance across every pair of layers 8--23 in
LLaVA-OneVision on ScanQA, complementing the layer-distance averages in
Section~\ref{sec:layer_dynamics}. As in the main analysis, a token's relevance
at each layer is its attention from the final prompt token, averaged over
attention heads. Figure~\ref{fig:app_ranking_heatmap}(a) measures the Spearman
correlation between the complete visual-token rankings at two layers.
Figure~\ref{fig:app_ranking_heatmap}(b) focuses on the highest-ranked tokens:
for each layer pair, it counts the visual tokens shared by their top-200 sets
and divides by 200.

Both heatmaps show the strongest agreement near the diagonal, with agreement
generally decreasing as the distance between layers increases. Adjacent layers
thus tend to preserve the overall token ranking and share many of the same
high-relevance tokens, whereas the composition of the top-200 set changes
more substantially across distant layers. Taken together, these observations
motivate recoverable group-wise skipping: low-priority visual tokens
temporarily skip computation, and all visual tokens are regrouped at later
anchors.

\appendixwrapfinish
\FloatBarrier

\endgroup

\FloatBarrier
\subsection{Complementarity of Projector-Level and LLM-Level Compression}
\label{app:two_stage}

\begin{wraptable}{R}{0.50\textwidth}
  \centering
  \vspace{-15pt}
  \caption{\small\textbf{Projector-only, LLM-only, and two-stage compression on LLaVA-OneVision.}
  Vis. tokens is the mean active count per decoder block; Avg. is retention
  relative to the dense model.}
  \label{tab:two_stage}
  \scriptsize
  \setlength{\tabcolsep}{2.5pt}
  \renewcommand{\arraystretch}{1.06}
  \resizebox{\linewidth}{!}{%
  \begin{tabular}{@{}lrr|rrr|r@{}}
    \toprule
    \multirow{2}{*}{\textbf{Configuration}}
      & \multirow{2}{*}{\makecell{\textbf{Proj.}\\\textbf{keep}}}
      & \multirow{2}{*}{\makecell{\textbf{Vis.}\\\textbf{tokens}}}
      & \multicolumn{2}{c}{\textbf{ScanQA}}
      & \textbf{SQA3D}
      & \multirow{2}{*}{\textbf{Avg. (\%)}} \\
    \cmidrule(lr){4-5}\cmidrule(lr){6-6}
      & & & \textbf{CIDEr} & \textbf{EM} & \textbf{EM} & \\
    \midrule
    GeoSemZip only & 10\% & 628 & 87.44 & 25.41 & 56.69 & 94.00 \\
    GroupRoute only & 100\% & 2128 & 88.48 & 26.07 & 55.84 & 94.65 \\
    \rowcolor{oursblue}
    GeoGR (both) & 30\% & 638 & 88.25 & 26.20 & 56.61 & \best{95.18} \\
    \bottomrule
  \end{tabular}%
  }
\end{wraptable}

Table~\ref{tab:two_stage} compares projector-only, LLM-only, and two-stage
compression on LLaVA-OneVision. GeoSemZip-only retains approximately 10\% of
projector tokens, which are processed by every decoder block. GroupRoute-only
keeps all 6,272 projector tokens and uses GeoGR's window $[8,23]$, anchors
$[9,13]$, and keep ratio $\kappa=0.5$. GeoGR combines a 30\% projector budget
with the same GroupRoute schedule.
GeoGR retains 95.18\% of dense performance with 638 active visual tokens per
decoder block. Compressing only at the projector to a similar active-token
count (628) lowers retention to 94.00\%, a 1.18-point difference. GroupRoute-only retains 94.65\% while processing 2,128 active tokens per block. These results support the effectiveness of combining spatial--semantic token compression before the LLM with query-conditioned group-wise skipping within the LLM.

\end{document}